\documentclass[10pt,twocolumn,letterpaper]{article}
\PassOptionsToPackage{table}{xcolor}
\usepackage[pagenumbers]{cvpr}
\def\confName{CVPR}
\def\confYear{2027}
\usepackage[T1]{fontenc}
\usepackage[utf8]{inputenc}
\usepackage{amsmath,amssymb,mathtools,bm}
\usepackage{graphicx}
\usepackage{booktabs,multirow,array}
\usepackage{algorithm}
\usepackage{algpseudocode}
\usepackage{xcolor}
\definecolor{cvprblue}{rgb}{0.21,0.49,0.74}
\usepackage[pagebackref,breaklinks,colorlinks,allcolors=cvprblue]{hyperref}
\usepackage{url}
\hypersetup{pdftitle={R2LR: Reliable Reward Learning for High-Fidelity Reference-Based Image Editing}}

\newcommand{\Method}{R$^{2}$LR}
\newcommand{\ScoreSet}{\{0,1,2,3,4,5\}}
\newcommand{\Dtrain}{\mathcal{D}_{\mathrm{train}}}
\newcommand{\Dh}{\mathcal{D}_{h}}

\newcommand{\Dall}{\mathcal{D}}
\newcommand{\RefDim}{\mathrm{ref}}
\newcommand{\AdhDim}{\mathrm{adh}}
\newcommand{\PreDim}{\mathrm{pre}}
\newcommand{\QualDim}{\mathrm{qual}}
\newcommand{\softmax}{\operatorname{softmax}}
\newcommand{\TopOp}{\operatorname{Top}}
\newcommand{\round}{\operatorname{round}}

\title{\texorpdfstring{$R^2$LR}{R2LR}: Reliable Reward Learning for High-Fidelity Reference-Based Image Editing}
\author{%
    Xijie Huang\textsuperscript{1,*}\quad Donghao Luo\textsuperscript{2,*}\quad Chao Wen\textsuperscript{2}\quad Chengming Xu\textsuperscript{2,\textdagger}\quad Peng TANG\textsuperscript{2}\quad Caoshuo Li\textsuperscript{2}\\[2pt]
    Xu Peng\textsuperscript{2}\quad Yongqi Yang\textsuperscript{2}\quad Xuyi Hu\textsuperscript{3}\quad Junyi He\textsuperscript{1}\quad Chengjie Wang\textsuperscript{2}\quad Yanwei Fu\textsuperscript{1,4,\textdagger}\\[2pt]
    {\normalsize \textsuperscript{1}Fudan University\quad \textsuperscript{2}Tencent YouTu Lab\quad \textsuperscript{3}University of Cambridge\quad \textsuperscript{4}Shanghai Innovation Institute}\\[2pt]
    {\small \textsuperscript{*}Equal contribution.\qquad \textsuperscript{\textdagger}Corresponding authors.}%
}
\hypersetup{pdfauthor={Xijie Huang}}

\begin{document}
\twocolumn[{%
    \renewcommand\twocolumn[1][]{#1}%
    \maketitle
    \begin{center}
        \captionsetup{type=figure,hypcap=false}
        \includegraphics[width=0.95\textwidth]{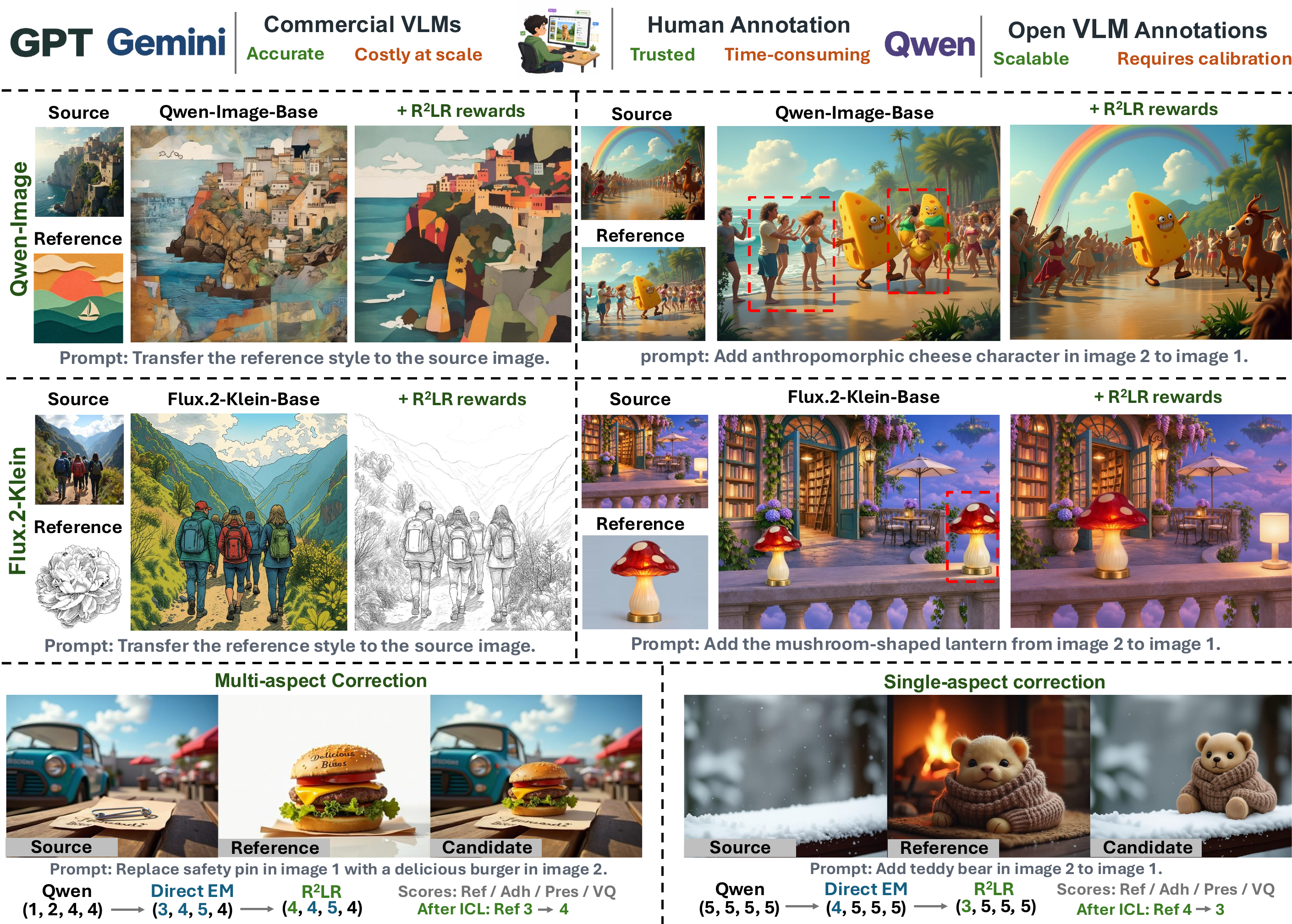}
        \vspace{-0.10in}
        \caption{Label repair and downstream editing with \Method{}. Top: reward-guided editing with Qwen-Image and FLUX.2-klein-4B. Bottom: multi- and single-aspect score corrections.}
        \label{fig:teaser}
        \vspace{-0.06in}
    \end{center}%
}]
\begin{abstract}
Reference-based image editing requires rewards that jointly evaluate reference consistency, instruction adherence, source preservation, and visual quality. Obtaining such four-aspect supervision at scale is difficult: human ratings and commercial vision--language models (VLMs) are costly, whereas open-source VLMs suffer from entangled systematic bias and uninformative guessing. We present \emph{Reliability-Guided Reward Label Repair} (\Method{}), which decouples whether an automatic score is informative (reliability) from how it should be calibrated (bias correction). Anchored by a modest set of human-scored examples, an expectation--maximization (EM) model learns an aspect-specific reliability gate and an instruction-conditioned confusion matrix. High-risk annotations undergo targeted in-context re-scoring with retrieved human demonstrations before filtering, and retained scores are calibrated via the informative-channel posterior. We then train a single-pass four-aspect reward model with \emph{reliability-weighted aspect-wise tie supervision} to constrain equal target aspect scores according to label reliability. \Method{} reduces annotation error, outperforms open-source VLMs and commercial judges (including GPT-5.6 and Gemini-3.1-pro) in human preference alignment, and improves downstream RL-tuned editing on Qwen-Image and FLUX.2-klein-4B.
\end{abstract}

\section{Introduction}

Reference-based image editing modifies a source image according to a language instruction and a visual exemplar~\citep{chen2024anydoor, ye2023ip, yangpaint}. Unlike text-only generation or editing~\citep{brooks2023instructpix2pix, hertz2022prompt, mokady2023nulltext}, it must transfer requested reference attributes—such as identity, appearance, or style—while keeping unrelated source regions intact. Since a visually plausible output may still copy the wrong attribute or alter the background, a faithful reward must jointly assess four aspects: \emph{reference consistency}, \emph{instruction adherence}, \emph{source preservation}, and \emph{visual quality}.

While reward models effectively align text-to-image generation~\citep{xu2024imagereward,pickscore,wuhps} and instruction-guided editing~\citep{wu2025editreward,luo2025editscore}, four-aspect reference-based editing faces a supervision bottleneck (Figure~\ref{fig:teaser}, top): human ratings are trustworthy but slow, commercial vision--language models (VLMs) such as GPT or Gemini are costly at scale, and open-source VLMs~\citep{bai2025qwen3} pose two coupled noise challenges. First, \emph{VLM errors mix systematic bias with uninformative guessing}. An annotator may grasp relative edit quality under a shifted scale, or fail on fine-grained cross-image comparison and guess blindly. In Figure~\ref{fig:teaser} (bottom), Qwen-VL overly penalizes a faithful burger replacement across aspects, yet assigns full marks \texttt{(5,5,5,5)} to a teddy bear addition with clear identity drift; confidence filtering cannot recalibrate biased scores, while post-hoc mapping cannot salvage guesses. Second, \emph{low-reliability annotations often occur on hard boundary cases that reward learning needs most}, such as subtle identity drift and local preservation errors, so naive filtering~\citep{han2018coteaching,northcutt2021confident} discards valuable supervision alongside noise.

Our key insight is to \emph{decouple whether an annotation is informative (reliability) from how its value should be calibrated (bias correction)}. We instantiate this in \textbf{\Method{}} (\emph{Reliability-Guided Reward Label Repair}), modeling each observed VLM score as a latent mixture of an \emph{informative channel} (structured confusions) and a \emph{guessing channel} (independent of true quality). Anchored by human-scored examples, an expectation--maximization (EM) procedure learns an aspect-specific reliability gate from token uncertainty and prompt semantics alongside a confusion model. Because scoring bias reflects both global annotator leniency and instruction-specific tendencies (e.g., local replacement vs.\ global style transfer), we parameterize the confusion matrix as a global base transition adapted by a lightweight instruction-conditioned low-rank perturbation.

Rather than discarding low-reliability annotations outright, \Method{} adopts a \emph{recovery-before-filtering} strategy. For the highest-risk annotations per aspect, we retrieve similar human-scored demonstrations for targeted in-context learning (ICL) re-scoring and reassess them with the frozen gate, removing a sample only if a targeted aspect remains unreliable. All retained automatic annotations—whether originally reliable or recovered via ICL—then undergo Bayesian posterior correction conditioned strictly on the informative channel, removing systematic bias without mixing in guessing noise.

To preserve these fine-grained labels during reward learning, we further address \emph{aspect-level ties that coarse overall rankings leave unconstrained}. Two edits often differ overall while sharing equal repaired scores on individual criteria (e.g., equal background preservation). Pairwise overall ranking ignores these equalities, whereas unweighted tie constraints ignore varying label certainty. We therefore train a shared VLM backbone to predict four-aspect Gaussian rewards in one forward pass, augmenting its overall ranking loss~\citep{wu2025editreward} with \emph{reliability-weighted aspect-wise tie supervision} that pulls predicted aspect means together according to joint label reliability.

Experiments show that \Method{} lowers label error, surpasses raw-label baselines, open-source VLMs, and commercial judges (including GPT-5.6 and Gemini-3.1-pro) in human preference alignment, and improves downstream RL-tuned editing on Qwen-Image and FLUX.2-klein-4B (Figure~\ref{fig:teaser}, top). Our contributions are threefold:
\begin{itemize}
    \setlength{\itemsep}{0pt}\setlength{\parskip}{0pt}\setlength{\topsep}{2pt}
    \item We propose \Method{}, a human-anchored label repair framework that disentangles biased informative scores from uninformative guesses for targeted ICL recovery, filtering, and reliable-channel posterior calibration.
    \item We introduce single-pass multi-dimensional reward learning with reliability-weighted aspect-wise tie supervision, enforcing aspect-level equality constraints according to label reliability.
    \item We show that \Method{} reduces label error, outperforms open-source and commercial VLM judges in four-aspect preference alignment, and improves downstream reference-based editing across backbones.
\end{itemize}

\section{Related Work}

\noindent \textbf{Reference-based and instruction-guided image editing.}
Text-guided editing manipulates images via cross-attention control, inversion, or instruction tuning~\citep{brooks2023instructpix2pix,hertz2022prompt,mokady2023nulltext}, whereas reference-based editing adds visual conditions for identity, appearance, or style via personalization~\citep{gal2023textualinversion,ruiz2023dreambooth,kumari} or exemplar adapters~\citep{yangpaint,ye2023ip,chen2024anydoor}. Evaluating these edits requires jointly assessing reference transfer, instruction adherence, source preservation, and visual quality—the four criteria targeted by our supervision.

\noindent \textbf{Reward models for visual generation and editing.}
ImageReward, PickScore, and HPS~\citep{xu2024imagereward,pickscore,wuhps} learn visual preferences for generation, while RewardDance~\citep{wu2025rewarddance} scales generative reward models against reward hacking. EditReward and EditScore~\citep{wu2025editreward,luo2025editscore} address instruction-guided editing without a visual exemplar or reference-consistency score. We adapt EditReward's Gaussian ranking formulation to four-aspect supervision and compare training objectives on the same repaired data, evaluating the training formulation rather than claiming superiority over full pretrained editing models.

\noindent \textbf{Learning from noisy and heterogeneous supervision.}
Latent-variable EM models estimate annotator reliability and hidden labels~\citep{dawid1979maximum,raykar2010learning}, alongside noise-transition, loss-correction, sample-selection, and confident-learning methods~\citep{goldberger2017training,patrini2017making,han2018coteaching,northcutt2021confident}. We extend these ideas to heterogeneous four-aspect reward annotations by pairing an informative confusion channel with a guessing channel, retrieving human examples to re-score uncertain cases, and calibrating retained scores.

\section{Reliable Reward Learning}
\label{sec:method}

\noindent \textbf{Problem Formulation and Overview}. 
Our method repairs automatic reward annotations before reward-model training.
Each editing sample is $x_i=(I_i^{\mathrm{src}},I_i^{\mathrm{ref}},
I_i^{\mathrm{edit}},p_i)$, containing the source, reference, edited image, and
instruction. Let $\Dall$ denote the full sample pool. We evaluate four aspects
$a\in\mathcal A=\{\RefDim,\AdhDim,\PreDim,\QualDim\}$:
reference consistency (Ref), instruction adherence (Adh), source preservation
(Pres), and visual quality (VQ). The latent clean score $y_i^a$ and observed VLM score $m_i^a$ lie in
$\ScoreSet$, with higher scores indicating better edits.
A set $\Dh\subset\Dall$ of 20K samples with human scores $y_{i,h}^a$ anchors this scale.
Our goal is to construct repaired scores $\hat y_i^a$ and accompanying
reliability estimates for reward learning.

\begin{figure*}
    \centering
    \includegraphics[width=0.95\linewidth]{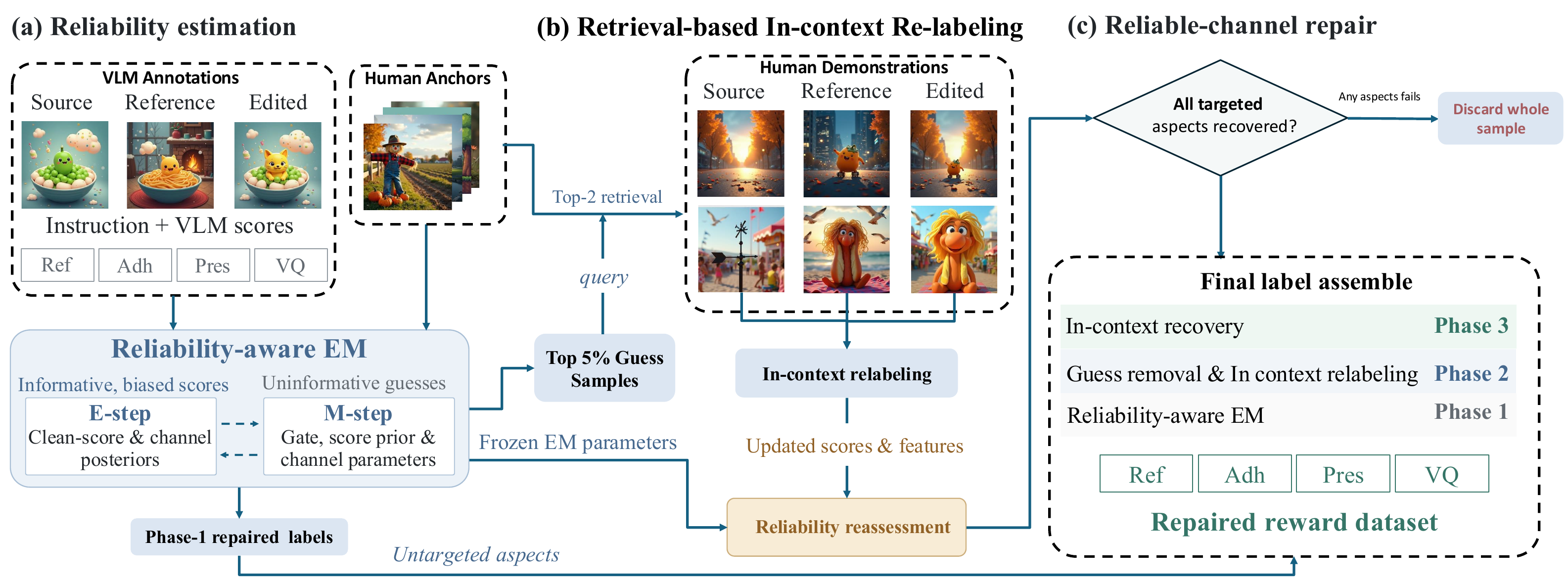}
    \vspace{-0.1in}
    \caption{Overview of \Method{}. EM estimates annotation reliability and score bias. Selected aspects are re-scored with human examples and accepted using updated reliability. All retained automatic scores, including aspects not selected for ICL, undergo posterior correction. The output contains four repaired scores and their associated reliabilities.    \label{fig:pipeline} }
    \vspace{-0.15in}

\end{figure*}

Figure~\ref{fig:pipeline} outlines three stages of \Method{}: estimate
aspect-specific annotation reliability and score bias, recover high-risk cases
through targeted in-context learning (ICL) for re-scoring, and correct retained labels through
the reliable-channel posterior. Human labels remain unchanged.

\subsection{Learning the Annotator's Reliable Regime}
VLM annotations vary in reliability across editing cases and aspects.
Even informative scores may be biased, motivating separate models of
score calibration and uninformative guessing.

\noindent\textbf{Two annotation channels.}
For each aspect, $z_i^a\in\{0,1\}$ indicates whether the annotator operates
in the informative, potentially biased channel. As a working model of
annotation noise, we write
\begin{equation}
\label{eq:mixture}
    P(m_i^a\mid y_i^a,x_i)
    =\rho_i^a T_i^a(y_i^a,m_i^a)+(1-\rho_i^a)G^a(m_i^a).
\end{equation}
Here $\rho_i^a=P(z_i^a=1\mid x_i)$ estimates annotation reliability from
the features defined below, $T_i^a(y,m)$ is a conditional confusion matrix,
and $G^a(m)$ is an
aspect-specific guessing distribution independent of the clean score.
The two channels distinguish informative but biased scores from
uninformative annotations; they are modeling assumptions rather than
observed internal states of the VLM.

\noindent\textbf{Reliability representation.}
We predict the gate from annotation uncertainty and editing semantics:
\begin{equation}
\label{eq:rho}
    \rho_i^a=\sigma(g_\phi^a(h_i^a)).
\end{equation}
The descriptor $h_i^a$ combines score-distribution entropy, top-score
log-probability margins, and a principal component analysis (PCA) projection of a T5 text embedding of the instruction
and aspect-specific scoring prompt; $\sigma$ denotes the logistic sigmoid.
Uncertainty features capture scoring confidence, while
semantic features describe the editing and evaluation requirements.
Features are normalized using training-set statistics.
Feature construction and regularization are detailed in the supplementary material.

\noindent\textbf{Instruction-conditioned score calibration.}
Even in the reliable channel, score bias can vary with the editing case:
localized replacement and fine-grained reference matching need not induce
the same score confusions. We adapt global logits
$A_0^a\in\mathbb R^{6\times6}$ through a low-rank perturbation:
\begin{equation}
\label{eq:lowrank}
\begin{aligned}
    A_i^a &= A_0^a+U^a\operatorname{diag}(c_i^a){V^a}^{\top},\\
    T_i^a(k,m) &= \softmax_m(A_i^a[k,m]),
\end{aligned}
\end{equation}
where $c_i^a=W^a e_i^a+b^a$, $e_i^a$ is the semantic feature, and
$U^a,V^a\in\mathbb R^{6\times r_{\mathrm{rank}}}$ are low-rank factors,
with $r_{\mathrm{rank}}=2$.
The row-wise softmax gives the observed-score distribution for each clean
score. The low-rank update limits the number of parameters learned from human
anchors. Because $e_i^a$ contains text features, candidates with the same
instruction share this matrix; their uncertainty features can still yield
different reliability gates. An ordinal regularizer favors nearby score
confusions.

\noindent\textbf{Estimation with human anchors.}
We jointly estimate annotation reliability and score bias with EM, using
$\pi_k^a=P(y_i^a=k)$ as the clean-score prior. For automatic annotations,
the E-step computes the joint posterior
$q_i^a(y,z)=P(y_i^a=y,z_i^a=z\mid m_i^a,x_i)$ under the current parameters.
For human anchors, we fix the clean score to the human label while keeping
the channel assignment latent, since their VLM annotations may still be
unreliable.

Holding these posteriors fixed, the M-step updates the prior, guessing
distribution, reliability gate, and conditional confusion parameters by
maximizing
\begin{equation}
\label{eq:em-objective}
\begin{aligned}
    \mathcal Q_{\mathrm{EM}}
    ={}&\sum_{i,a}\mathbb E_{q_i^a(y,z)}
    \Bigl[\log\pi_y^a\\
    &\quad+z\log\!\bigl(\rho_i^a T_i^a(y,m_i^a)\bigr)\\
    &\quad+(1-z)\log\!\bigl((1-\rho_i^a)G^a(m_i^a)\bigr)\Bigr]\\
    &-\Omega_{\mathrm{ord}}-\Omega_{\mathrm{reg}},
\end{aligned}
\end{equation}
where $\Omega_{\mathrm{ord}}$ favors nearby score confusions and
$\Omega_{\mathrm{reg}}$ collects the remaining regularization terms.
The posterior responsibility $r_i^a=\sum_{k=0}^{5}q_i^a(k,1)$ also
conditions on the observed score, distinguishing it from the gate
$\rho_i^a$ used for selection and reward learning.
We perform three EM iterations in Phase~1 and keep the fitted parameters
fixed during subsequent ICL relabeling and score correction.
Full EM updates and regularization details are provided in the
supplementary material.

\subsection{Budgeted Aspect Relabeling}
Difficult editing cases may still provide valuable supervision.
We therefore attempt to recover high-risk annotations before filtering.
For each aspect, we select
\begin{equation}
\label{eq:highrisk}
    \mathcal H^a=\TopOp_{\alpha}(1-\rho_i^a),
\end{equation}
where $\TopOp_\alpha$ returns the indices of the highest-risk $\alpha=5\%$
of automatic samples in that aspect. The four selected sets can overlap.
Samples selected in any aspect are temporarily excluded from the training
pool pending ICL recovery.
Selection uses $\rho_i^a$, since compatibility of an observed score with
the fitted model alone does not establish that the annotation is informative.

We retrieve two human-labeled demonstrations using DINOv2~\citep{dinov2}
and OpenCLIP~\citep{openclip} features of the source--reference--edited tuple,
with optional instruction features.
Each demonstration includes the three images, instruction, four human scores,
and brief scoring explanations. All four scores provide evaluation context,
while relabeling targets only the selected aspect. Retrieval, descriptor updates, and acceptance thresholds are detailed
in the supplementary material.

ICL changes the scoring prompt without updating the annotator's parameters.
It returns a new observation $\tilde m_i^a$ and an updated descriptor
$\tilde h_i^a$. With the Phase~1 EM parameters fixed, we recompute the
aspect-wise reliability $\tilde\rho_i^a=\sigma(g_\phi^a(\tilde h_i^a))$.
An aspect is recovered only if $\tilde\rho_i^a>\tau_a$, where $\tau_a$ is
its acceptance threshold. Let
$\mathcal U^a=\{i\in\mathcal H^a:\tilde\rho_i^a\leq\tau_a\}$.
We discard the entire automatic sample if any selected aspect fails this
criterion or if its final four-aspect labels are incomplete. Denoting the
latter set by $\mathcal U_{\mathrm{miss}}$, the final removal set is
\begin{equation}
\label{eq:remove}
    \mathcal U=\mathcal U_{\mathrm{miss}}\cup
    \bigcup_{a\in\mathcal A}\mathcal U^a.
\end{equation}
A sample returns to the training pool only when all its selected aspects
are recovered and complete four-aspect supervision is available.
For retained samples, the accepted ICL scores are then corrected using
the same EM model, as described below.

\subsection{Bias Correction in the Reliable Regime}
Reliability screening determines which annotations to retain; it does not
remove their residual score bias. For retained automatic annotations, we
use the same Phase~1 EM model, without refitting, to infer the clean score
through the reliable channel. Let
$m_i^{a,*}=\tilde m_i^a$ if the aspect was accepted after ICL relabeling,
and $m_i^{a,*}=m_i^a$ otherwise. The repair posterior is
conditioned on $z_i^a=1$, and its mean is rounded to the ordinal scale:
\begin{equation}
\label{eq:posterior}
\begin{aligned}
    &P(y_i^a=k\mid m_i^{a,*},z_i^a=1,x_i)\\
    &\qquad=\frac{\pi_k^a T_i^a(k,m_i^{a,*})}
    {\sum_{k'=0}^{5}\pi_{k'}^a T_i^a(k',m_i^{a,*})},
\end{aligned}
\end{equation}
\begin{equation}
\label{eq:posterior-repair}
    \hat y_i^a=\round\!\left(
    \sum_{k=0}^{5}k\,P(y_i^a=k\mid m_i^{a,*},z_i^a=1,x_i)\right).
\end{equation}
Conditioning on $z_i^a=1$ excludes the guessing channel, which helps
assess reliability but provides no clean-score information for correction.
Human anchors retain $\hat y_i^a=y_{i,h}^a$. The repaired training pool is
\begin{equation}
\label{eq:dtrain}
    \Dtrain=\Dh\cup\left((\Dall\setminus\Dh)\setminus\mathcal U\right).
\end{equation}
Each retained automatic score carries reliability $\rho_i^{a,*}$ from the
observation used for correction: $\tilde\rho_i^a$ after accepted ICL and
$\rho_i^a$ otherwise. Human labels have training reliability
$\rho_i^{a,*}=1$; this does not fix the latent channel of their VLM annotations
during EM. The complete label-repair algorithm is provided in the supplementary material.

\section{Reward Learning and Editing Optimization}
\label{sec:reward-learning}

\subsection{Efficient Multi-dimensional Reward Model}

A shared VLM backbone reads the source, reference, and edited images
together with the instruction. A joint scoring prompt and four
aspect-specific tokens enable prediction of all four aspects in one
forward pass. Following EditReward's Gaussian
parameterization~\citep{wu2025editreward}, head $a$ outputs a mean $\mu_i^a$
and log standard deviation $\ell_i^a$, defining
$P_{\theta,i}^a=\mathcal N(\mu_i^a,(\sigma_i^a)^2)$, where
$\sigma_i^a=\exp(\ell_i^a)$. The mean $R_\theta^a(x_i)=\mu_i^a$ serves as
the aspect reward. Candidates in a pair are scored independently.

\noindent \textbf{Reward ranking objective.}
We retain EditReward's uncertainty-aware ranking
objective~\citep{wu2025editreward} as the base loss.
Let $s_i^a=\hat y_i^a$ denote the training label.
For a pair $(A,B)$ sharing an editing condition, the overall preference is
$d_{AB}=\operatorname{sign}(\sum_a s_A^a-\sum_a s_B^a)$.
Following EditReward, we average the predicted means and log standard
deviations across aspects and sample the reward difference:
\begin{equation}
\label{eq:base-sampling}
\begin{aligned}
 \bar\mu_i&=\tfrac14\sum_a\mu_i^a,\qquad
 \bar\ell_i=\tfrac14\sum_a\ell_i^a,\\
 z_{AB}^{(k)}&=\bar\mu_A-\bar\mu_B+
 \sqrt{e^{2\bar\ell_A}+e^{2\bar\ell_B}}\,\epsilon_k,
 \quad\epsilon_k\sim\mathcal N(0,1).
\end{aligned}
\end{equation}
With $K=1{,}000$ Monte Carlo samples, define
$f_{AB}=K^{-1}\sum_k\operatorname{softplus}(-z_{AB}^{(k)})$ and
$b_{AB}=K^{-1}\sum_k\operatorname{softplus}(z_{AB}^{(k)})$.
For a minibatch $\mathcal B$, the base ranking objective is
\begin{equation}
\label{eq:base-ranking}
 \mathcal L_{\mathrm{base}}=
 \frac{c}{|\mathcal B|}\sum_{(A,B)\in\mathcal B}
 \begin{cases}
 f_{AB},&d_{AB}>0,\\
 b_{AB},&d_{AB}<0,\\
 f_{AB}+b_{AB},&d_{AB}=0.
 \end{cases}
\end{equation}
Here, $c$ is a fixed normalization constant.
Overall ties contribute both preference directions within the same
training pair.

\noindent \textbf{Reliability-weighted aspect ties.}
Our four-aspect labels contain equal-score relations that overall
preference supervision does not explicitly constrain.
Expanding these aspect-level ties into opposing binary preferences would
introduce additional training records.
We encode these relations directly through an aspect-wise tie penalty.
Since equal scores can have different annotation reliabilities, we weight
each constraint by the reliability of both labels.

Let $\rho_i^{a,*}$ denote the training reliability,
$t_{AB}^a=\mathbf1[s_A^a=s_B^a]$,
$w_{AB}^a=\sqrt{\rho_A^{a,*}\rho_B^{a,*}}$, and
$\Delta_{AB}^a=\mu_A^a-\mu_B^a$. We define
\begin{equation}
\label{eq:mu-tie}
\begin{aligned}
 g(u)&=\tfrac12[\operatorname{softplus}(u)
                  +\operatorname{softplus}(-u)]-\log2,\\
 \mathcal L_{\mathrm{tie}}&=
 \frac{1}{4|\mathcal B|}\sum_{(A,B)\in\mathcal B}\sum_a
 t_{AB}^a w_{AB}^a g(\Delta_{AB}^a).
\end{aligned}
\end{equation}
The penalty is minimized when the two predicted aspect means are equal,
with stronger supervision for more reliable ties.
We average the weighted penalties over all $4|\mathcal B|$ aspect pairs.
The final objective is
\begin{equation}
\label{eq:reward-loss}
    \mathcal L_R=\mathcal L_{\mathrm{base}}
    +\lambda_{\mathrm{tie}}\mathcal L_{\mathrm{tie}},
    \qquad\lambda_{\mathrm{tie}}=0.5.
\end{equation}
Normalization details and the binary-tie baseline are described in the
supplementary material.

\subsection{Downstream Reward-Guided Optimization}

We adopt the Edit-R1 training framework~\citep{edit-r1}, which is
based on DiffusionNFT~\citep{diffusionnft}, and replace its original reward
model with our learned reward model. The four predicted aspect means are
combined into a scalar reward:
\begin{equation}
\label{eq:scalar-reward}
\begin{aligned}
    R(x)&=w_{\RefDim}R_\theta^{\RefDim}(x)+w_{\AdhDim}R_\theta^{\AdhDim}(x)\\
    &\quad+w_{\PreDim}R_\theta^{\PreDim}(x)+w_{\QualDim}R_\theta^{\QualDim}(x),
\end{aligned}
\end{equation}
where all weights are fixed at $0.25$.
We retain the DiffusionNFT training objective and use $R(x)$ as reward
feedback. Further training details are provided in the supplementary material.

\begin{figure*}
    \centering
    \includegraphics[width=\linewidth]{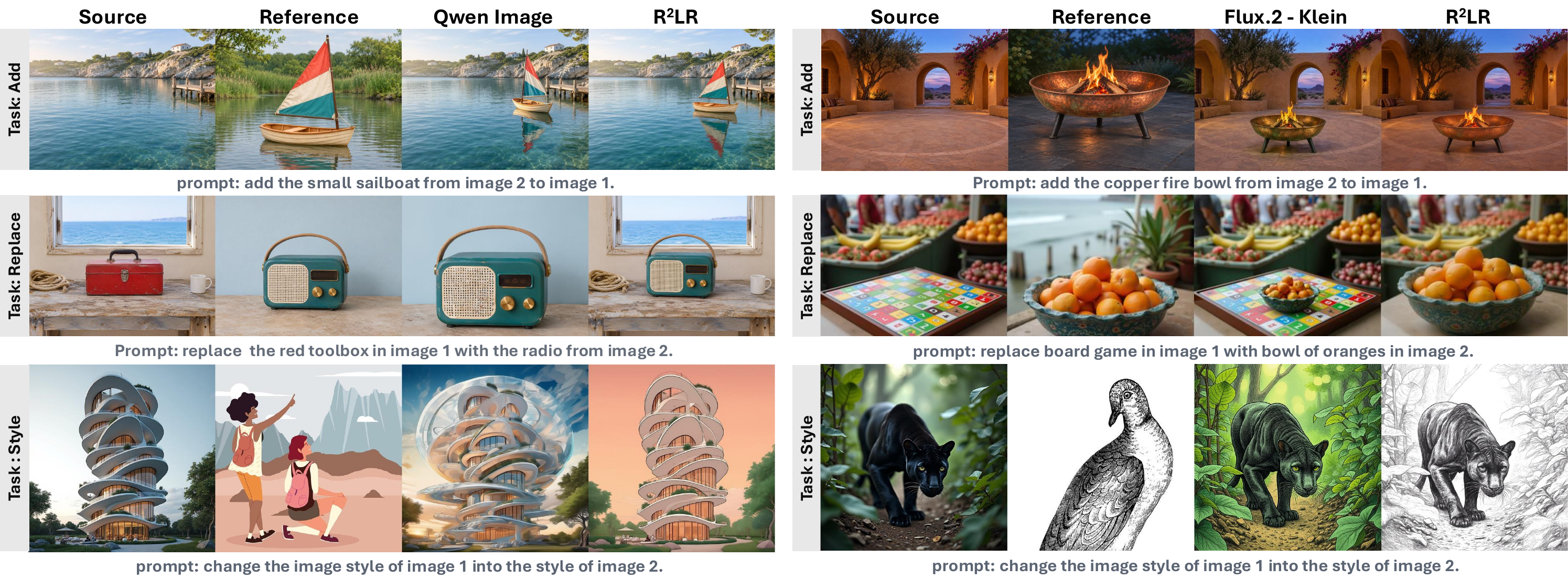}
    \vspace{-0.1in}
    \caption{Qualitative results with base models on addition, replacement, and style transfer. Left: Qwen-Image; right: FLUX.2-klein-4B.}
    \label{fig:main_results}
    \vspace{-0.15in}
\end{figure*}

\section{Experiments}

\begin{table*}[t]
\centering\small
\caption{\textbf{Reward model comparison} (accuracy).
$\geq k$ denotes agreement on at least $k$ aspects ($k=2,3$);
All-4 and All-NT require agreement on all four and all human non-tie aspects, respectively. \textbf{Bold} and \underline{underline} mark the best and second-best results.}
\label{tab:rm-main}
\begingroup
\setlength{\tabcolsep}{2pt}
\setlength{\arrayrulewidth}{0.30pt}
\setlength{\aboverulesep}{0pt}
\setlength{\belowrulesep}{0pt}
\renewcommand{\arraystretch}{1.08}
\resizebox{\linewidth}{!}{%
\begin{tabular}{@{}l|rrrr|r|rrr|rrrr|r|rrr@{}}
\toprule
\multirow{3}{*}{\textbf{Method}} & \multicolumn{8}{c|}{\textbf{Tie-aware}} & \multicolumn{8}{c}{\textbf{Strict non-tie}} \\
\cline{2-17}
 & \multicolumn{4}{c|}{Aspect-wise} & \multirow{2}{*}{Mean} & \multicolumn{3}{c|}{Joint agreement} & \multicolumn{4}{c|}{Aspect-wise} & \multirow{2}{*}{Mean} & \multicolumn{3}{c}{Joint agreement} \\
\cline{2-5}\cline{7-13}\cline{15-17}
 & Ref & Adh & Pres & VQ & & $\geq2$ & $\geq3$ & All-4 & Ref & Adh & Pres & VQ & & $\geq2$ & $\geq3$ & All-NT \\
\midrule
\multicolumn{17}{c}{\textit{\textbf{Zero-shot open-source VLMs}}} \\
\hline
Qwen2.5-VL-7B & 0.351 & 0.340 & 0.378 & 0.444 & 0.378 & 0.451 & 0.213 & 0.087 & 0.232 & 0.166 & 0.205 & 0.112 & 0.178 & 0.154 & 0.079 & 0.082 \\
Qwen3-VL-4B & 0.286 & 0.331 & 0.352 & 0.453 & 0.356 & 0.423 & 0.214 & 0.085 & 0.133 & 0.139 & 0.108 & 0.158 & 0.134 & 0.129 & 0.091 & 0.059 \\
Qwen3-VL-8B & 0.312 & 0.333 & 0.354 & 0.435 & 0.359 & 0.421 & 0.206 & 0.073 & 0.178 & 0.139 & 0.113 & 0.094 & 0.131 & 0.096 & 0.046 & 0.043 \\
Qwen3-VL-32B & 0.345 & 0.366 & 0.420 & 0.462 & 0.398 & 0.495 & 0.228 & 0.067 & 0.219 & 0.234 & 0.247 & 0.273 & 0.243 & 0.219 & 0.109 & 0.084 \\
MiMo-VL-7B & 0.320 & 0.358 & 0.398 & 0.413 & 0.372 & 0.455 & 0.223 & 0.097 & 0.199 & 0.218 & 0.254 & 0.196 & 0.217 & 0.225 & 0.143 & 0.124 \\
\midrule
\multicolumn{17}{c}{\textit{\textbf{Few-shot open-source VLMs}}} \\
\hline
Qwen3-VL-8B (2-shot) & 0.253 & 0.293 & 0.348 & 0.430 & 0.331 & 0.392 & 0.193 & 0.072 & 0.046 & 0.047 & 0.087 & 0.065 & 0.061 & 0.063 & 0.027 & 0.030 \\
Qwen3-VL-32B (2-shot) & 0.341 & 0.385 & 0.399 & 0.449 & 0.394 & 0.491 & 0.248 & 0.094 & 0.227 & 0.229 & 0.201 & 0.135 & 0.198 & 0.211 & 0.117 & 0.095 \\
\midrule
\multicolumn{17}{c}{\textit{\textbf{Commercial VLMs}}} \\
\hline
GPT-5.6-Luna & 0.457 & 0.506 & 0.644 & 0.496 & 0.526 & 0.686 & 0.374 & 0.119 & 0.468 & 0.523 & 0.739 & 0.356 & 0.521 & 0.539 & 0.330 & 0.260 \\
GPT-5.6-Sol & 0.453 & 0.522 & \textbf{0.701} & 0.457 & 0.533 & 0.722 & 0.363 & 0.123 & 0.469 & 0.526 & \underline{0.817} & 0.446 & \underline{0.564} & \underline{0.606} & 0.330 & \underline{0.281} \\
GPT-5.6-Terra & \underline{0.471} & 0.507 & 0.698 & 0.455 & 0.533 & 0.692 & 0.386 & \underline{0.135} & \underline{0.484} & 0.484 & 0.771 & \underline{0.473} & 0.553 & 0.572 & \underline{0.353} & 0.278 \\
Gemini-3.1-pro & 0.432 & \underline{0.554} & \underline{0.700} & \textbf{0.511} & \underline{0.549} & \underline{0.728} & \underline{0.421} & 0.113 & 0.437 & \underline{0.551} & 0.716 & 0.427 & 0.533 & 0.581 & 0.300 & 0.229 \\
\midrule
\multicolumn{17}{c}{\textit{\textbf{Reward models}}} \\
\hline
\rowcolor{gray!15}
\textbf{\Method{}} & \textbf{0.583} & \textbf{0.623} & 0.672 & \underline{0.497} & \textbf{0.594} & \textbf{0.753} & \textbf{0.494} & \textbf{0.213} & \textbf{0.743} & \textbf{0.811} & \textbf{0.840} & \textbf{0.694} & \textbf{0.772} & \textbf{0.810} & \textbf{0.679} & \textbf{0.577} \\
\bottomrule
\end{tabular}%
}
\endgroup
\vspace{-0.1in}
\end{table*}

\begin{table}[t]
\caption{\textbf{Label-repair and reward-supervision ablations on the evaluation set}
(accuracy). All models use Qwen3-VL-8B. Bold marks the best result.}
\label{tab:rm-ablation}
\centering\footnotesize
\begingroup
\setlength{\tabcolsep}{1.2pt}
\setlength{\arrayrulewidth}{0.30pt}
\setlength{\aboverulesep}{0pt}
\setlength{\belowrulesep}{0pt}
\renewcommand{\arraystretch}{1.10}
\resizebox{\columnwidth}{!}{%
\begin{tabular}{@{}l|rrrr|r|rrr@{}}
\toprule
\multicolumn{9}{c}{\textbf{Tie-aware}} \\
\midrule
\multirow{2}{*}{\textbf{Method}} & \multicolumn{4}{c|}{Aspect-wise} & \multirow{2}{*}{Mean} & \multicolumn{3}{c}{Joint agreement} \\
\cline{2-5}\cline{7-9}
 & Ref & Adh & Pres & VQ & & $\geq2$ & $\geq3$ & All-4 \\
\midrule
Binary-tie & 0.557 & \textbf{0.634} & 0.642 & 0.482 & 0.579 & 0.752 & 0.466 & 0.190 \\
Qwen Anno & 0.561 & 0.586 & 0.572 & 0.458 & 0.544 & 0.685 & 0.438 & 0.160 \\
Direct EM & 0.563 & 0.611 & 0.670 & 0.495 & 0.585 & 0.747 & 0.485 & 0.189 \\
w/o ICL & 0.570 & 0.612 & 0.671 & 0.496 & 0.587 & 0.752 & 0.486 & 0.197 \\
\textbf{\Method{}} & \textbf{0.583} & 0.623 & \textbf{0.672} & \textbf{0.497} & \textbf{0.594} & \textbf{0.753} & \textbf{0.494} & \textbf{0.213} \\
\midrule
\multicolumn{9}{c}{\textbf{Strict non-tie}} \\
\midrule
\multirow{2}{*}{\textbf{Method}} & \multicolumn{4}{c|}{Aspect-wise} & \multirow{2}{*}{Mean} & \multicolumn{3}{c}{Joint agreement} \\
\cline{2-5}\cline{7-9}
 & Ref & Adh & Pres & VQ & & $\geq2$ & $\geq3$ & All-NT \\
\midrule
Binary-tie & 0.713 & 0.809 & 0.828 & 0.653 & 0.751 & 0.791 & 0.639 & 0.538 \\
Qwen Anno & 0.717 & 0.758 & 0.763 & 0.625 & 0.716 & 0.753 & 0.612 & 0.530 \\
Direct EM & 0.710 & 0.793 & 0.831 & 0.690 & 0.756 & 0.795 & 0.651 & 0.547 \\
w/o ICL & 0.731 & 0.794 & 0.833 & 0.692 & 0.763 & 0.799 & 0.662 & 0.572 \\
\textbf{\Method{}} & \textbf{0.743} & \textbf{0.811} & \textbf{0.840} & \textbf{0.694} & \textbf{0.772} & \textbf{0.810} & \textbf{0.679} & \textbf{0.577} \\
\bottomrule
\end{tabular}%
}
\endgroup
\vspace{-0.1in}
\end{table}

\begin{figure*}
    \centering
    \includegraphics[width=\linewidth]{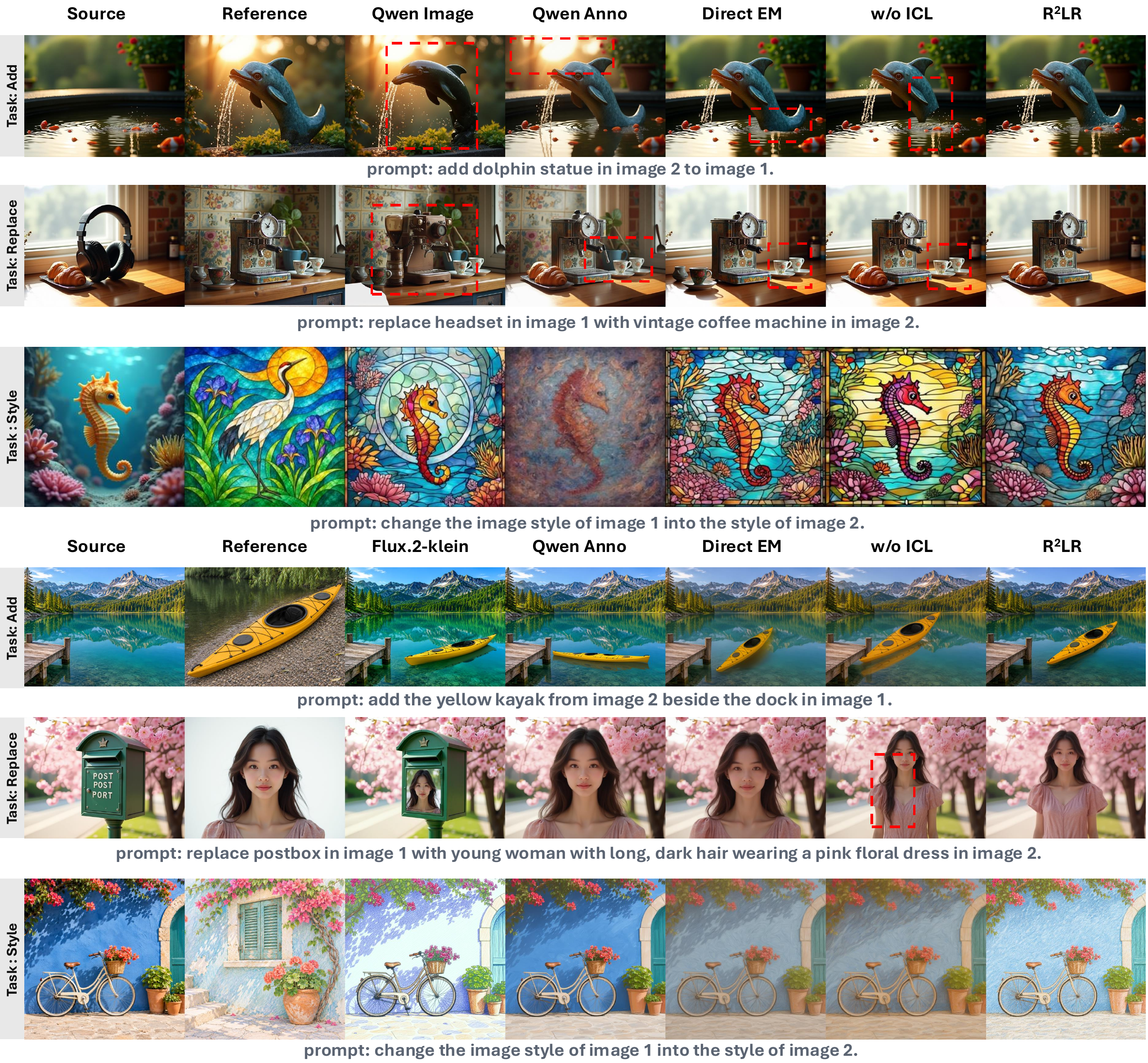}
    \vspace{-0.12in}
    \caption{Qualitative comparison of reward supervision variants on Qwen-Image (top) and FLUX.2-klein-4B (bottom). Red boxes highlight editing errors.    \label{fig:ablation-qual} }
    \vspace{-0.18in}

\end{figure*}
\noindent \textbf{Setup}.
Following~\citep{pengcheng}, we construct $50{,}000$ source--reference--instruction seeds and generate $300{,}000$ candidates with six open-source editors: DreamOmni2~\citep{dreamomni2}, OmniGen2~\citep{omnigen2}, Flux2~\citep{flux-2-2025}, Echo-4o~\citep{echo-4o}, and Qwen-Image~\citep{qwenimage}. Qwen3-VL-235B-A22B-Instruct supplies initial four-aspect scores, anchored by $20{,}000$ human-scored samples. From each supervision pool, we build $250$K same-condition pairs to fine-tune Qwen3-VL-8B-Instruct for one epoch, and evaluate reward alignment on $1{,}000$ held-out human-scored pairs disjoint from all training sets (protocols and costs in the supplementary material).

\subsection{Aligning Rewards with Human Preferences}
\label{sec:rm-alignment}
Each model independently scores two candidate edits across four aspects against human judgments (Table~\ref{tab:rm-main}). Following EditReward~\citep{wu2025editreward}, continuous reward differences use aspect-specific tie thresholds, whereas integer VLM scores use exact equality. We evaluate tie-aware accuracy over all pairs, strict non-tie accuracy on pairs with unequal human scores, and joint agreement requiring correctness across multiple aspects simultaneously (at least two, at least three, all four aspects, or all human non-tie aspects).

Table~\ref{tab:rm-main} highlights three findings. First, zero-shot open-source VLMs fail to separate fine-grained edit quality: Qwen3-VL-32B reaches $0.398$ tie-aware accuracy by frequently outputting identical template scores that coincide with human ties, but drops to $0.243$ on strict non-ties. Second, naive test-time few-shot prompting hurts discrimination: two-shot prompting lowers strict non-tie accuracy on Qwen3-VL-8B ($0.131\rightarrow0.061$) and Qwen3-VL-32B ($0.243\rightarrow0.198$) by biasing outputs toward the demonstration scores. Third, \Method{} ($0.594$ tie-aware, $0.772$ non-tie) substantially outperforms both open-source VLMs and commercial judges. While GPT-5.6 variants and Gemini-3.1-pro perform well on source preservation, they struggle with cross-image reference consistency ($0.437$--$0.484$ non-tie) and instruction adherence ($0.484$--$0.551$). \Method{} lifts strict non-tie accuracy to $0.743$ on reference consistency and $0.811$ on instruction adherence, more than doubling joint accuracy across all non-tie aspects ($0.577$ vs.\ $0.281$ for the strongest commercial judge). We further examine whether improved alignment with human preferences translates into better downstream editing results.

\subsection{Downstream Multi-Image Editing}
\label{sec:downstream}
We next use each learned reward model to guide DiffusionNFT fine-tuning of Qwen-Image~\citep{qwenimage} and FLUX.2-klein-4B~\citep{flux-2-2025}, scored by GPT-5.6-Terra on a $0$--$5$ scale (Table~\ref{tab:downstream}; MixGRPO~\citep{mixgrpo} results in the supplementary material). \Method{} raises the overall score from $2.9162$ to $4.3705$ on Qwen-Image and from $3.6862$ to $4.1262$ on FLUX.2-klein-4B. Examining individual aspects shows that base Qwen-Image struggles most with source preservation ($2.3225$) and reference consistency ($2.7550$). Fine-tuning with raw VLM supervision brings limited gain on preservation ($3.0175$), whereas \Method{} drives preservation to $4.6660$ and reference consistency to $4.1640$. In Figure~\ref{fig:main_results}, \Method{} enables faithful attribute transfer—such as cleanly replacing the board game with a bowl of oranges or applying sketch style without background hallucinations—while keeping unedited regions intact.

\begin{table}[t]
\caption{\textbf{Downstream editing quality}, evaluated by GPT-5.6-Terra
(scores $0$--$5$). Bold marks the best per column.}
\label{tab:downstream}
\footnotesize
\centering
\begingroup
\setlength{\tabcolsep}{3pt}
\renewcommand{\arraystretch}{1.10}
\begin{tabular*}{\columnwidth}{@{\extracolsep{\fill}}l rrrrr@{}}
\toprule
\multicolumn{6}{c}{\textbf{Qwen-Image}} \\
\midrule
\textbf{Method} & Ref & Adh & Pres & VQ & Overall \\
\midrule
Base
  & 2.7550 & 3.0150 & 2.3225 & 3.5725 & 2.9162 \\
Qwen Anno
  & 3.6400 & 3.6900 & 3.0175 & 3.8500 & 3.5494 \\
Direct EM
  & 3.8025 & 3.7200 & 3.8800 & \textbf{4.3700} & 3.9431 \\
w/o ICL
  & 4.1540 & 4.3600 & 4.6320 & 4.1960 & 4.3355 \\
\textbf{\Method{}}
  & \textbf{4.1640} & \textbf{4.4060} & \textbf{4.6660} & 4.2460 & \textbf{4.3705} \\
\midrule
\multicolumn{6}{c}{\textbf{FLUX.2-klein-4B}} \\
\midrule
\textbf{Method} & Ref & Adh & Pres & VQ & Overall \\
\midrule
Base
  & 3.3575 & 3.5875 & 3.8850 & 3.9150 & 3.6862 \\
Qwen Anno
  & 3.7950 & 4.0750 & 3.9750 & 4.0350 & 3.9700 \\
Direct EM
  & 3.7800 & 4.0625 & 4.2900 & 3.9525 & 4.0213 \\
w/o ICL
  & 3.8850 & 4.1150 & 4.1825 & 4.0125 & 4.0488 \\
\textbf{\Method{}}
  & \textbf{3.9175} & \textbf{4.1475} & \textbf{4.3750} & \textbf{4.0650} & \textbf{4.1262} \\
\bottomrule
\end{tabular*}
\endgroup

\end{table}

\subsection{Ablation Study}
\label{sec:ablation}
We ablate label repair and reward supervision in Tables~\ref{tab:rm-ablation}--\ref{tab:downstream} and Figure~\ref{fig:ablation-qual} using the Qwen3-VL-8B-Instruct backbone across four supervision stages: raw VLM annotations ($300$K samples), EM posterior calibration on the full pool without filtering ($300$K), reliability filtering of the lowest-$5\%$ scores per aspect without in-context recovery ($250{,}688$ samples), and our full \Method{} pipeline ($274{,}939$ samples).

\noindent \textbf{EM calibration and reliability filtering.}
In Table~\ref{tab:rm-ablation}, applying EM posterior calibration alone raises mean non-tie accuracy from $0.716$ to $0.756$, led by source preservation ($0.763\rightarrow0.831$) and visual quality ($0.625\rightarrow0.690$) where systematic scale bias dominates (reducing held-out label MAE from $1.168$ to $0.991$). Yet calibration alone slightly lowers reference consistency ($0.717\rightarrow0.710$), as posterior mapping cannot salvage uninformative cross-image guesses. Filtering out low-reliability annotations (whose raw MAE is $1.780$ vs.\ $0.635$ for high-reliability samples) rebounds reference consistency to $0.731$ and raises mean non-tie accuracy to $0.763$.

\noindent \textbf{Targeted ICL recovery.}
Recovering $24{,}251$ hard cases via human-guided in-context re-scoring in \Method{} further lifts strict non-tie accuracy to $0.743$ on reference consistency and $0.811$ on instruction adherence, while improving full four-aspect agreement from $0.197$ to $0.213$. Visually, Figure~\ref{fig:ablation-qual} shows that editors trained without guess filtering suffer from style and structure artifacts, whereas filtering without recovery misses fine reference details (such as the coffee cups and dress texture in red boxes) that \Method{} resolves.

\noindent \textbf{Reliability-weighted tie supervision.}
Compared with EditReward's binary tie expansion trained on the exact same repaired labels, our reliability-weighted aspect-wise tie objective improves tie-aware ($0.579\rightarrow0.594$), strict non-tie ($0.751\rightarrow0.772$), and joint non-tie ($0.538\rightarrow0.577$) accuracy by explicitly constraining single-aspect equalities according to label reliability.

\subsection{User Study}
\label{sec:user-study}
Twenty participants each rate $10$ randomly sampled editing cases across all five variants on a $0$--$5$ scale (Table~\ref{tab:user-study}). While unfiltered calibration and filtering without recovery mainly improve human ratings on source preservation ($3.222$ and $3.378$ vs.\ $2.556$ for raw VLM supervision), \Method{} achieves the highest ratings across every criterion, substantially boosting reference consistency ($3.056\rightarrow3.822$), instruction adherence ($2.833\rightarrow3.778$), and overall quality ($2.992\rightarrow3.775$).

\begin{table}[htbp] \small
\centering
\caption{User study on editing quality (mean human ratings).
\textbf{Bold} marks the best per column.}
\label{tab:user-study}
\setlength{\tabcolsep}{4pt}
\renewcommand{\arraystretch}{1.0}
\begin{tabular}{lrrrrr}
\toprule
\textbf{Method} & Ref & Adh & Pres & VQ & Overall \\
\midrule
Base & 2.022 & 2.067 & 2.178 & 2.322 & 2.147 \\
Qwen Anno & 2.689 & 2.667 & 2.556 & 2.633 & 2.636 \\
Direct EM & 2.800 & 2.856 & 3.222 & 2.689 & 2.892 \\
w/o ICL & 3.056 & 2.833 & 3.378 & 2.700 & 2.992 \\
\Method{} & \textbf{3.822} & \textbf{3.778}
& \textbf{3.900} & \textbf{3.600} & \textbf{3.775} \\
\bottomrule
\end{tabular}
\vspace{-0.1in}
\end{table}

\section{Conclusion}

We presented \Method{}, a framework for improving automatic reward
annotations using human anchors. By separating annotation informativeness from score calibration, \Method{} guides targeted re-scoring to recover reliable supervision for difficult cases and corrects retained scores through the informative channel. Reliability-weighted aspect-wise tie supervision uses the repaired scores and their reliabilities to train a four-aspect reward model. Experiments demonstrate stronger agreement with human preferences in four-aspect evaluation and improved editing quality on two editing backbones, supporting reliable label repair for scalable reward learning from noisy automatic annotations.

\section*{LLM Usage Statement}
ChatGPT and Codex were used to assist with manuscript
polishing, including improvements to grammar, wording, and
clarity. They also provided assistance with coding and
debugging, analysis of experimental results, and discussion
of research ideas.

{\small
\bibliographystyle{ieeenat_fullname}
\bibliography{sample-bibliography}}

\begin{thebibliography}{33}
\providecommand{\natexlab}[1]{#1}
\providecommand{\url}[1]{\texttt{#1}}
\expandafter\ifx\csname urlstyle\endcsname\relax
  \providecommand{\doi}[1]{doi: #1}\else
  \providecommand{\doi}{doi: \begingroup \urlstyle{rm}\Url}\fi

\bibitem[Bai et~al.(2025)Bai, Cai, Chen, Chen, Chen, Cheng, Deng, Ding, Gao,
  Ge, et~al.]{bai2025qwen3}
Shuai Bai, Yuxuan Cai, Ruizhe Chen, Keqin Chen, Xionghui Chen, Zesen Cheng,
  Lianghao Deng, Wei Ding, Chang Gao, Chunjiang Ge, et~al.
\newblock Qwen3-vl technical report.
\newblock \emph{arXiv preprint arXiv:2511.21631}, 2025.

\bibitem[Brooks et~al.(2023)Brooks, Holynski, and
  Efros]{brooks2023instructpix2pix}
Tim Brooks, Aleksander Holynski, and Alexei~A. Efros.
\newblock Instructpix2pix: Learning to follow image editing instructions.
\newblock In \emph{Proceedings of the IEEE/CVF Conference on Computer Vision
  and Pattern Recognition}, pages 18392--18402, 2023.

\bibitem[Chen et~al.(2024)Chen, Huang, Liu, Shen, Zhao, and
  Zhao]{chen2024anydoor}
Xi Chen, Lianghua Huang, Yu Liu, Yujun Shen, Deli Zhao, and Hengshuang Zhao.
\newblock Anydoor: Zero-shot object-level image customization.
\newblock In \emph{Proceedings of the IEEE/CVF Conference on Computer Vision
  and Pattern Recognition}, 2024.

\bibitem[Dawid and Skene(1979)]{dawid1979maximum}
Alexander~Philip Dawid and Allan~M. Skene.
\newblock Maximum likelihood estimation of observer error-rates using the em
  algorithm.
\newblock \emph{Journal of the Royal Statistical Society: Series C (Applied
  Statistics)}, 28\penalty0 (1):\penalty0 20--28, 1979.

\bibitem[Gal et~al.(2023)Gal, Alaluf, Atzmon, Patashnik, Bermano, Chechik, and
  Cohen-Or]{gal2023textualinversion}
Rinon Gal, Yuval Alaluf, Yuval Atzmon, Or Patashnik, Amit~H. Bermano, Gal
  Chechik, and Daniel Cohen-Or.
\newblock An image is worth one word: Personalizing text-to-image generation
  using textual inversion.
\newblock In \emph{International Conference on Learning Representations}, 2023.

\bibitem[Goldberger and Ben-Reuven(2017)]{goldberger2017training}
Jacob Goldberger and Ehud Ben-Reuven.
\newblock Training deep neural-networks using a noise adaptation layer.
\newblock In \emph{International Conference on Learning Representations}, 2017.

\bibitem[Han et~al.(2018)Han, Yao, Yu, Niu, Xu, Hu, Tsang, and
  Sugiyama]{han2018coteaching}
Bo Han, Quanming Yao, Xingrui Yu, Gang Niu, Miao Xu, Weihua Hu, Ivor Tsang, and
  Masashi Sugiyama.
\newblock Co-teaching: Robust training of deep neural networks with extremely
  noisy labels.
\newblock In \emph{Advances in Neural Information Processing Systems}, 2018.

\bibitem[Hertz et~al.(2022)Hertz, Mokady, Tenenbaum, Aberman, Pritch, and
  Cohen-Or]{hertz2022prompt}
Amir Hertz, Ron Mokady, Jay Tenenbaum, Kfir Aberman, Yael Pritch, and Daniel
  Cohen-Or.
\newblock Prompt-to-prompt image editing with cross attention control.
\newblock \emph{arXiv preprint arXiv:2208.01626}, 2022.

\bibitem[Ilharco et~al.(2021)Ilharco, Wortsman, Carlini, Taori, Dave, Shankar,
  Namkoong, Miller, Hajishirzi, Farhadi, et~al.]{openclip}
Gabriel Ilharco, Mitchell Wortsman, Nicholas Carlini, Rohan Taori, Achal Dave,
  Vaishaal Shankar, Hongseok Namkoong, John Miller, Hannaneh Hajishirzi, Ali
  Farhadi, et~al.
\newblock Openclip.
\newblock \emph{Zenodo}, 2021.

\bibitem[Kirstain et~al.(2023)Kirstain, Polyak, Singer, Matiana, Penna, and
  Levy]{pickscore}
Yuval Kirstain, Adam Polyak, Uriel Singer, Shahbuland Matiana, Joe Penna, and
  Omer Levy.
\newblock Pick-a-pic: An open dataset of user preferences for text-to-image
  generation.
\newblock \emph{Advances in neural information processing systems},
  36:\penalty0 36652--36663, 2023.

\bibitem[Kumari et~al.(2023)Kumari, Zhang, Zhang, Shechtman, and Zhu]{kumari}
Nupur Kumari, Bingliang Zhang, Richard Zhang, Eli Shechtman, and Jun-Yan Zhu.
\newblock Multi-concept customization of text-to-image diffusion.
\newblock In \emph{Proceedings of the IEEE/CVF conference on computer vision
  and pattern recognition}, pages 1931--1941, 2023.

\bibitem[Labs(2025)]{flux-2-2025}
Black~Forest Labs.
\newblock {FLUX.2: Frontier Visual Intelligence}.
\newblock \url{https://bfl.ai/blog/flux-2}, 2025.

\bibitem[Li et~al.(2025{\natexlab{a}})Li, Cui, Huang, Ma, Fan, Yang, and
  Zhong]{mixgrpo}
Junzhe Li, Yutao Cui, Tao Huang, Yinping Ma, Chun Fan, Miles Yang, and Zhao
  Zhong.
\newblock Mixgrpo: Unlocking flow-based grpo efficiency with mixed ode-sde,
  2025{\natexlab{a}}.

\bibitem[Li et~al.(2025{\natexlab{b}})Li, Liu, Zhang, Lin, Yuan, Yan, Ye, Yu,
  Niu, and Yuan]{edit-r1}
Zongjian Li, Zheyuan Liu, Qihui Zhang, Bin Lin, Shenghai Yuan, Zhiyuan Yan,
  Yang Ye, Wangbo Yu, Yuwei Niu, and Li Yuan.
\newblock Uniworld-v2: Reinforce image editing with diffusion negative-aware
  finetuning and mllm implicit feedback.
\newblock \emph{arXiv preprint arXiv:2510.16888}, 2025{\natexlab{b}}.

\bibitem[Luo et~al.(2025)Luo, Wang, Wu, Xiao, Jiang, Lian, Zhang, and
  Liu]{luo2025editscore}
Xin Luo, Jiahao Wang, Chenyuan Wu, Shitao Xiao, Xiyan Jiang, Defu Lian, Jiajun
  Zhang, and Zheng Liu.
\newblock Editscore: Unlocking online rl for image editing via high-fidelity
  reward modeling.
\newblock \emph{arXiv preprint arXiv:2509.23909}, 2025.

\bibitem[Mokady et~al.(2023)Mokady, Hertz, Aberman, Pritch, and
  Cohen-Or]{mokady2023nulltext}
Ron Mokady, Amir Hertz, Kfir Aberman, Yael Pritch, and Daniel Cohen-Or.
\newblock Null-text inversion for editing real images using guided diffusion
  models.
\newblock In \emph{Proceedings of the IEEE/CVF Conference on Computer Vision
  and Pattern Recognition}, pages 6038--6047, 2023.

\bibitem[Northcutt et~al.(2021)Northcutt, Jiang, and
  Chuang]{northcutt2021confident}
Curtis~G. Northcutt, Lu Jiang, and Isaac~L. Chuang.
\newblock Confident learning: Estimating uncertainty in dataset labels.
\newblock \emph{Journal of Artificial Intelligence Research}, 70:\penalty0
  1373--1411, 2021.

\bibitem[Oquab et~al.(2023)Oquab, Darcet, Moutakanni, Vo, Szafraniec, Khalidov,
  Fernandez, Haziza, Massa, El-Nouby, et~al.]{dinov2}
Maxime Oquab, Timoth{\'e}e Darcet, Th{\'e}o Moutakanni, Huy Vo, Marc
  Szafraniec, Vasil Khalidov, Pierre Fernandez, Daniel Haziza, Francisco Massa,
  Alaaeldin El-Nouby, et~al.
\newblock Dinov2: Learning robust visual features without supervision.
\newblock \emph{arXiv preprint arXiv:2304.07193}, 2023.

\bibitem[Patrini et~al.(2017)Patrini, Rozza, Krishna~Menon, Nock, and
  Qu]{patrini2017making}
Giorgio Patrini, Alessandro Rozza, Aditya Krishna~Menon, Richard Nock, and
  Lizhen Qu.
\newblock Making deep neural networks robust to label noise: A loss correction
  approach.
\newblock In \emph{Proceedings of the IEEE Conference on Computer Vision and
  Pattern Recognition}, pages 1944--1952, 2017.

\bibitem[Raykar et~al.(2010)Raykar, Yu, Zhao, Valadez, Florin, Bogoni, and
  Moy]{raykar2010learning}
Vikas~C. Raykar, Shipeng Yu, Linda~H. Zhao, Gerardo~Hermosillo Valadez, Charles
  Florin, Luca Bogoni, and Linda Moy.
\newblock Learning from crowds.
\newblock \emph{Journal of Machine Learning Research}, 11:\penalty0 1297--1322,
  2010.

\bibitem[Ruiz et~al.(2023)Ruiz, Li, Jampani, Pritch, Rubinstein, and
  Aberman]{ruiz2023dreambooth}
Nataniel Ruiz, Yuanzhen Li, Varun Jampani, Yael Pritch, Michael Rubinstein, and
  Kfir Aberman.
\newblock Dreambooth: Fine tuning text-to-image diffusion models for
  subject-driven generation.
\newblock In \emph{Proceedings of the IEEE/CVF Conference on Computer Vision
  and Pattern Recognition}, pages 22500--22510, 2023.

\bibitem[Wu et~al.(2025{\natexlab{a}})Wu, Li, Zhou, Lin, Gao, Yan, ming Yin,
  Bai, Xu, Chen, Chen, Tang, Zhang, Wang, Yang, Yu, Cheng, Liu, Li, Zhang,
  Meng, Wei, Ni, Chen, Cao, Peng, Qu, Wu, Wang, Yu, Wen, Feng, Xu, Wang, Zhang,
  Zhu, Wu, Cai, and Liu]{qwenimage}
Chenfei Wu, Jiahao Li, Jingren Zhou, Junyang Lin, Kaiyuan Gao, Kun Yan, Sheng
  ming Yin, Shuai Bai, Xiao Xu, Yilei Chen, Yuxiang Chen, Zecheng Tang, Zekai
  Zhang, Zhengyi Wang, An Yang, Bowen Yu, Chen Cheng, Dayiheng Liu, Deqing Li,
  Hang Zhang, Hao Meng, Hu Wei, Jingyuan Ni, Kai Chen, Kuan Cao, Liang Peng,
  Lin Qu, Minggang Wu, Peng Wang, Shuting Yu, Tingkun Wen, Wensen Feng,
  Xiaoxiao Xu, Yi Wang, Yichang Zhang, Yongqiang Zhu, Yujia Wu, Yuxuan Cai, and
  Zenan Liu.
\newblock Qwen-image technical report, 2025{\natexlab{a}}.

\bibitem[Wu et~al.(2025{\natexlab{b}})Wu, Zheng, Yan, Xiao, Luo, Wang, Li,
  Jiang, Liu, Zhou, et~al.]{omnigen2}
Chenyuan Wu, Pengfei Zheng, Ruiran Yan, Shitao Xiao, Xin Luo, Yueze Wang, Wanli
  Li, Xiyan Jiang, Yexin Liu, Junjie Zhou, et~al.
\newblock Omnigen2: Exploration to advanced multimodal generation.
\newblock \emph{arXiv preprint arXiv:2506.18871}, 2025{\natexlab{b}}.

\bibitem[Wu et~al.(2025{\natexlab{c}})Wu, Gao, Ye, Li, Li, Guo, Liu, Xue, Hou,
  Liu, Zeng, and Huang]{wu2025rewarddance}
Jie Wu, Yu Gao, Zilyu Ye, Ming Li, Liang Li, Hanzhong Guo, Jie Liu, Zeyue Xue,
  Xiaoxia Hou, Wei Liu, Yan Zeng, and Weilin Huang.
\newblock Rewarddance: Reward scaling in visual generation.
\newblock \emph{arXiv preprint arXiv:2509.08826}, 2025{\natexlab{c}}.

\bibitem[Wu et~al.(2025{\natexlab{d}})Wu, Jiang, Ku, Nie, Liu, and
  Chen]{wu2025editreward}
Keming Wu, Sicong Jiang, Max Ku, Ping Nie, Minghao Liu, and Wenhu Chen.
\newblock Editreward: A human-aligned reward model for instruction-guided image
  editing.
\newblock \emph{arXiv preprint arXiv:2509.26346}, 2025{\natexlab{d}}.

\bibitem[Wu et~al.(2023)Wu, Sun, Zhu, Zhao, and Li]{wuhps}
Xiaoshi Wu, Keqiang Sun, Feng Zhu, Rui Zhao, and Hongsheng Li.
\newblock Human preference score: Better aligning text-to-image models with
  human preference.
\newblock In \emph{Proceedings of the IEEE/CVF International Conference on
  Computer Vision}, pages 2096--2105, 2023.

\bibitem[Xia et~al.(2025)Xia, Peng, Zhang, Huang, Liu, Li, Tan, Wu, Wang, Wang,
  et~al.]{dreamomni2}
Bin Xia, Bohao Peng, Yuechen Zhang, Junjia Huang, Jiyang Liu, Jingyao Li, Haoru
  Tan, Sitong Wu, Chengyao Wang, Yitong Wang, et~al.
\newblock Dreamomni2: Multimodal instruction-based editing and generation.
\newblock \emph{arXiv preprint arXiv:2510.06679}, 2025.

\bibitem[Xu et~al.(2024)Xu, Liu, Wu, Tong, Li, Ding, Tang, and
  Dong]{xu2024imagereward}
Jiazheng Xu, Xiao Liu, Yuchen Wu, Yuxuan Tong, Qinkai Li, Ming Ding, Jie Tang,
  and Yuxiao Dong.
\newblock Imagereward: Learning and evaluating human preferences for
  text-to-image generation.
\newblock In \emph{Advances in Neural Information Processing Systems}, 2024.

\bibitem[Xu et~al.(2026)Xu, Tang, Luo, Hu, Cui, He, Chen, Zhang, Ling, and
  Wang]{pengcheng}
Pengcheng Xu, Peng Tang, Donghao Luo, Xiaobin Hu, Weichu Cui, Qingdong He,
  Zhennan Chen, Jiangning Zhang, Charles Ling, and Boyu Wang.
\newblock Towards generalized multi-image editing for unified multimodal
  models.
\newblock \emph{arXiv preprint arXiv:2601.05572}, 2026.

\bibitem[Yang et~al.(2023)Yang, Gu, Zhang, Zhang, Chen, Sun, Chen, and
  Wen]{yangpaint}
Binxin Yang, Shuyang Gu, Bo Zhang, Ting Zhang, Xuejin Chen, Xiaoyan Sun, Dong
  Chen, and Fang Wen.
\newblock Paint by example: Exemplar-based image editing with diffusion models.
\newblock In \emph{Proceedings of the IEEE/CVF conference on computer vision
  and pattern recognition}, pages 18381--18391, 2023.

\bibitem[Ye et~al.(2023)Ye, Zhang, Liu, Han, and Yang]{ye2023ip}
Hu Ye, Jun Zhang, Sibo Liu, Xiao Han, and Wei Yang.
\newblock Ip-adapter: Text compatible image prompt adapter for text-to-image
  diffusion models.
\newblock \emph{arXiv preprint arXiv:2308.06721}, 2023.

\bibitem[Ye et~al.(2025)Ye, Jiang, Wang, Zhu, Hu, Huang, He, Yan, Yu, Li,
  et~al.]{echo-4o}
Junyan Ye, Dongzhi Jiang, Zihao Wang, Leqi Zhu, Zhenghao Hu, Zilong Huang, Jun
  He, Zhiyuan Yan, Jinghua Yu, Hongsheng Li, et~al.
\newblock Echo-4o: Harnessing the power of gpt-4o synthetic images for improved
  image generation.
\newblock \emph{arXiv preprint arXiv:2508.09987}, 2025.

\bibitem[Zheng et~al.(2025)Zheng, Chen, Ye, Wang, Zhang, Jiang, Su, Ermon, Zhu,
  and Liu]{diffusionnft}
Kaiwen Zheng, Huayu Chen, Haotian Ye, Haoxiang Wang, Qinsheng Zhang, Kai Jiang,
  Hang Su, Stefano Ermon, Jun Zhu, and Ming-Yu Liu.
\newblock Diffusionnft: Online diffusion reinforcement with forward process.
\newblock \emph{arXiv preprint arXiv:2509.16117}, 2025.

\end{thebibliography}
\end{document}


\setcounter{page}{1}
\maketitlesupplementary
\setcounter{figure}{4}
\setcounter{table}{4}
\renewcommand{\theequation}{S\arabic{equation}}
\renewcommand{\theHequation}{supp.\arabic{equation}}

\let\linkedEqref\eqref
\renewcommand{\eqref}[1]{%
  \begingroup
  \edef\eqtarget{\getrefbykeydefault{#1}{anchor}{}}%
  \ifx\eqtarget\empty
    \textup{(\textcolor{cvprblue}{\ref*{#1}})}%
  \else
    \linkedEqref{#1}%
  \fi
  \endgroup}
\makeatletter
\def\hyper@natlinkstart#1{\begingroup\color{cvprblue}}
\def\hyper@natlinkend{\endgroup}
\def\hyper@natlinkbreak#1#2{\endgroup#1\begingroup\color{cvprblue}}
\makeatother

\section{Additional Method Details}

\subsection{Annotation Model Parameterization}
\label{sec:annotation-details}

\noindent\textbf{Reliability features.}
For each aspect $a$, let $\ell_{\mathrm{ann},i}^a(k)$ denote the
annotator's log-probability for score token $k\in\ScoreSet$.
We normalize over these six tokens:
\begin{equation}
    \eta_i^a(k)=
    \frac{\exp(\ell_{\mathrm{ann},i}^a(k))}
    {\sum_{k'\in\ScoreSet}\exp(\ell_{\mathrm{ann},i}^a(k'))}.
\end{equation}
Let $b_{i,1},\ldots,b_{i,B_{\mathrm{tok}}}$ index the top
$B_{\mathrm{tok}}$ score tokens in descending probability.
The uncertainty descriptor combines normalized entropy and score margins:
\begin{equation}
\begin{aligned}
    u_i^a=\Big[
    &-\frac{1}{\log 6}\sum_{k\in\ScoreSet}
        \eta_i^a(k)\log\eta_i^a(k),\\
    &\ell_{\mathrm{ann},i}^a(b_{i,1})
        -\ell_{\mathrm{ann},i}^a(b_{i,2}),\\
    &\{\ell_{\mathrm{ann},i}^a(b_{i,j})
        -\ell_{\mathrm{ann},i}^a(b_{i,1})\}_{j=1}^{B_{\mathrm{tok}}}
    \Big].
\end{aligned}
\end{equation}
These features describe scoring uncertainty, while a semantic descriptor
captures the editing instruction and aspect-specific criteria. The gate uses
\begin{equation}
\begin{aligned}
    e_{i,\mathrm{gate}}^a&=\operatorname{PCA}\bigl(\operatorname{T5}([p_i;\psi^a])\bigr),\\
    h_i^a&=[\operatorname{Norm}(u_i^a),\operatorname{Norm}(e_{i,\mathrm{gate}}^a)].
\end{aligned}
\end{equation}
Here $\psi^a$ is the aspect-specific scoring prompt.
Feature normalization uses statistics from the training data, which remain
fixed during subsequent relabeling and evaluation.
The gate $\rho_i^a=\sigma(g_\phi^a(h_i^a))$ combines uncertainty and semantics.
The confusion matrix in Eq.~\eqref{eq:lowrank} is conditioned on the
instruction representation $e_i^a$.
Candidates sharing the same instruction therefore share this matrix, while
their image-dependent scoring uncertainty can yield different reliabilities.

\noindent\textbf{Ordinal regularization.}
We encourage nearby score confusions through
\begin{equation}
\label{eq:ordinal-reg}
    \Omega_{\mathrm{ord}}
    =\lambda_{\mathrm{ord}}\sum_{i,a}\sum_{k,m}
    T_i^a(k,m)(k-m)^2.
\end{equation}
The penalty increases with the distance between the clean and observed
scores, while allowing systematic shifts within the informative channel.
Marginalizing the clean score gives
\begin{equation}
\label{eq:marginal}
\begin{aligned}
    P(m_i^a\mid x_i)
    =\sum_{k=0}^{5}\pi_k^a
    \bigl[\rho_i^a T_i^a(k,m_i^a)+(1-\rho_i^a)G^a(m_i^a)\bigr].
\end{aligned}
\end{equation}

\begin{algorithm*}[t]
\caption{Reliability-Guided Reward Label Repair}
\label{alg:repair}
\begin{algorithmic}[1]
\Require Candidate pool $\Dall$, human anchors $\Dh$ with scores $y_{i,h}^a$, initial VLM scores $m_i^a$,
features $h_i^a$, aspects $\mathcal A$
\Ensure Repaired training records $(x_i,\{\hat y_i^a,\rho_i^{a,*}\}_{a\in\mathcal A})$
\State $\mathcal D_{\mathrm{auto}}\gets\Dall\setminus\Dh$
\Statex \textbf{Phase 1: Fit the annotation model}
\For{each $a\in\mathcal A$}
    \State Initialize the annotation model for aspect $a$
    \For{$t=1$ to $3$}
        \State E-step: compute $q^{a,1},q^{a,0}$ using
        Eqs.~\eqref{eq:q1}--\eqref{eq:human-q0}
        \State M-step: update the prior, guessing distribution, gate, and confusion model
        using Eq.~\eqref{eq:mstep}
    \EndFor
    \State Freeze the fitted parameters and compute $\rho_i^a$
    for $i\in\mathcal D_{\mathrm{auto}}$
\EndFor
\Statex \textbf{Phase 2: Recover selected aspects with ICL}
\For{each $a\in\mathcal A$}
    \State Initialize $m_i^{a,*}\gets m_i^a$, $\rho_i^{a,*}\gets\rho_i^a$,
    and $\mathcal U^a\gets\varnothing$
    \State Select the bottom $5\%$ by $\rho_i^a$ among automatic annotations as
    $\mathcal H^a$ and record the cutoff $\tau_a$
    \For{each $i\in\mathcal H^a$}
        \State Retrieve the two most visually similar human-scored demonstrations
        and re-score aspect $a$ to obtain $\tilde m_i^a,\tilde h_i^a$
        \State Compute $\tilde\rho_i^a=\sigma(g_\phi^a(\tilde h_i^a))$
        using the frozen Phase-1 gate
        \If{the re-scored annotation is valid and $\tilde\rho_i^a>\tau_a$}
            \State $(m_i^{a,*},\rho_i^{a,*})\gets(\tilde m_i^a,\tilde\rho_i^a)$
        \Else
            \State $\mathcal U^a\gets\mathcal U^a\cup\{i\}$
        \EndIf
    \EndFor
\EndFor
\State $\mathcal U_{\mathrm{miss}}\gets
\{i\in\mathcal D_{\mathrm{auto}}:\text{four-aspect supervision is incomplete}\}$
\State $\mathcal U\gets\mathcal U_{\mathrm{miss}}\cup\bigcup_{a\in\mathcal A}\mathcal U^a$;
$\mathcal D_{\mathrm{keep}}\gets\mathcal D_{\mathrm{auto}}\setminus\mathcal U$
\Statex \textbf{Phase 3: Correct retained scores with the frozen model}
\For{each $i\in\mathcal D_{\mathrm{keep}}$ and $a\in\mathcal A$}
    \State Set $\hat y_i^a$ to the rounded informative-channel posterior mean
    for $m_i^{a,*}$ using Eqs.~\eqref{eq:posterior}--\eqref{eq:posterior-repair}
\EndFor
\State For every human anchor, set $\hat y_i^a\gets y_{i,h}^a$ and $\rho_i^{a,*}\gets1$
\State $\Dtrain\gets\Dh\cup\mathcal D_{\mathrm{keep}}$
\State \Return $\{(x_i,\{\hat y_i^a,\rho_i^{a,*}\}_{a\in\mathcal A}):x_i\in\Dtrain\}$
\end{algorithmic}
\end{algorithm*}

\subsection{EM Estimation}
\label{sec:em-details}

For each automatic annotation, the E-step computes the joint posterior
over the clean score $y_i^a$ and channel $z_i^a$. Writing
$Z_i^a=P(m_i^a\mid x_i)$ from Eq.~\eqref{eq:marginal}, the responsibilities are
\begin{align}
\label{eq:q1}
    q_{i,k}^{a,1}
    &=P(y_i^a=k,z_i^a=1\mid m_i^a,x_i)
      =\frac{\pi_k^a\rho_i^a T_i^a(k,m_i^a)}{Z_i^a},\\
\label{eq:q0}
    q_{i,k}^{a,0}
    &=P(y_i^a=k,z_i^a=0\mid m_i^a,x_i)
      =\frac{\pi_k^a(1-\rho_i^a)G^a(m_i^a)}{Z_i^a}.
\end{align}
For a human anchor, the clean score is fixed to $y_{i,h}^a=k_h$, while
the channel remains latent:
\begin{align}
\label{eq:human-q}
    q_{i,k}^{a,1}
    &=\mathbf1[k=k_h]\,
      \frac{\rho_i^aT_i^a(k_h,m_i^a)}
      {\rho_i^aT_i^a(k_h,m_i^a)+(1-\rho_i^a)G^a(m_i^a)},\\
\label{eq:human-q0}
    q_{i,k}^{a,0}
    &=\mathbf1[k=k_h]\,
      \frac{(1-\rho_i^a)G^a(m_i^a)}
      {\rho_i^aT_i^a(k_h,m_i^a)+(1-\rho_i^a)G^a(m_i^a)}.
\end{align}
The informative-channel responsibility is
$r_i^a=\sum_kq_{i,k}^{a,1}$.
Unlike the feature-based gate $\rho_i^a$, this posterior also conditions
on the observed score. It supplies the soft channel assignment for fitting.

Holding the responsibilities fixed, the M-step maximizes
\begin{align}
\label{eq:mstep}
    \mathcal Q_{\mathrm{EM}}
    ={}&\sum_{i,a,k}q_{i,k}^{a,1}
    [\log\pi_k^a+\log\rho_i^a+\log T_i^a(k,m_i^a)]\nonumber\\
    &+\sum_{i,a,k}q_{i,k}^{a,0}
    [\log\pi_k^a+\log(1-\rho_i^a)+\log G^a(m_i^a)]\nonumber\\
    &-\Omega_{\mathrm{ord}}-\Omega_{\mathrm{reg}}.
\end{align}
Here $\Omega_{\mathrm{ord}}$ is defined in Eq.~\eqref{eq:ordinal-reg}.
$\Omega_{\mathrm{reg}}$ combines a squared $\ell_2$ penalty on the learned
gate and confusion parameters with the negative log densities of symmetric
Dirichlet priors on $\pi^a$ and $G^a$.
The categorical updates use expected clean-score and guessing-channel counts:
\begin{align}
    n_k^a&=\sum_i(q_{i,k}^{a,1}+q_{i,k}^{a,0}),\\
    v_m^a&=\sum_i\mathbf1[m_i^a=m]\sum_k q_{i,k}^{a,0}.
\end{align}
Normalizing and smoothing these counts updates $\pi^a$ and $G^a$,
respectively. Human anchors contribute hard clean-score counts to $\pi^a$
and responsibility-weighted observations to the confusion model, with
anchor weight $\gamma=1$.
The gate and conditional confusion parameters are updated by gradient steps
on Eq.~\eqref{eq:mstep}.

\noindent\textbf{Initialization and fitting.}
We initialize $\pi^a$ from the human-score histogram and use a rank-2
confusion update for each aspect. After three EM iterations, we freeze
the fitted model for ICL recovery and posterior correction.

\subsection{Targeted Relabeling and Posterior Correction}
\label{sec:relabel-details}

\noindent\textbf{Demonstration retrieval.}
For each image, we normalize its DINOv2~\citep{dinov2} and
OpenCLIP~\citep{openclip} features separately. We concatenate these
features in source--reference--edit order, normalize the resulting vector,
and retrieve the two human-scored demonstrations with the highest cosine
similarity to the query.

\noindent\textbf{Scoring context.}
Each demonstration provides the source, reference, and edited images,
the editing instruction, and four human scores. For each selected aspect
of the query, the annotator returns an integer from 0 to 5 followed by
a brief explanation grounded in the query images.
Sec.~\ref{sec:scoring-prompts} gives the image order and response format.

\noindent\textbf{Recovery by the reliability gate.}
For each aspect $a$, we retain the Phase-1 gate cutoff $\tau_a$ that defines
the lowest $5\%$ of automatic annotations. After ICL, we recompute the
uncertainty features from the new score-token probabilities and apply the
frozen Phase-1 normalization and gate to the updated descriptor $\tilde h_i^a$.
We accept a valid re-score only if
$\tilde\rho_i^a=\sigma(g_\phi^a(\tilde h_i^a))>\tau_a$.
A candidate is retained only when every targeted aspect passes and all four
scores are available.

\noindent\textbf{Final score correction.}
For a retained candidate, $m_i^{a,*}$ is the accepted ICL score for a
re-scored aspect and the original score for any other aspect.
All four automatic scores are corrected through the informative-channel
posterior in Eqs.~\eqref{eq:posterior}--\eqref{eq:posterior-repair}.
This step reuses the Phase-1 prior and confusion model without another EM fit.
Conditioning on $z_i^a=1$ cancels $\rho_i^{a,*}$ from the normalized
posterior, while its value remains available to weight reward supervision.
Human scores are retained unchanged and have training reliability 1.
Algorithm~\ref{alg:repair} summarizes the complete procedure.

\section{Label-Repair Diagnostics}
\label{sec:label-diagnostics}

\subsection{Candidate-Level Score Correction}
\label{sec:heldout-mae}

We apply the frozen Direct EM
model to the original Qwen scores, using the informative-channel posterior
without ICL or filtering.
Table~\ref{tab:label-diagnostics} compares the original and corrected
integer scores against the human ratings. Mean absolute error (MAE)
decreases on every aspect, with the largest reduction in visual quality.

\begin{table}[!htb]
\centering\small
\caption{Score correction on $1{,}902$ held-out candidates.
Mean averages the four aspect MAEs against the human scores.}
\label{tab:label-diagnostics}
\begin{tabular*}{\columnwidth}{@{\extracolsep{\fill}}lrr@{}}
\toprule
Aspect & Raw Qwen $\downarrow$ & EM-corrected $\downarrow$\\
\midrule
Ref & 1.3812 & \textbf{1.2161}\\
Adh & 1.1998 & \textbf{1.1104}\\
Pres & 0.9816 & \textbf{0.8638}\\
VQ & 1.1099 & \textbf{0.7744}\\
\midrule
Mean & 1.1681 & \textbf{0.9912}\\
\bottomrule
\end{tabular*}
\end{table}

\subsection{Reliability and Raw Annotation Error}
\label{sec:reliability-groups}

Using the $1{,}902$ edits from Sec.~\ref{sec:heldout-mae}, we rank
edits separately within each aspect by their pre-ICL reliability
$\rho_i^a$. The lowest and highest 100 per aspect give 400
candidate--aspect observations per group, spanning 358 and 315 distinct
edits, respectively. Raw Qwen scores have higher MAE in the low-$\rho$
group on all four aspects (Table~\ref{tab:reliability-diagnostics}).

\begin{table}[htb]
\centering\small
\caption{Raw Qwen MAE against human scores by pre-ICL reliability
($\downarrow$). Each group contains 100 edits per aspect.}
\label{tab:reliability-diagnostics}
\setlength{\tabcolsep}{3pt}
\begin{tabular*}{\columnwidth}{@{\extracolsep{\fill}}lrrrrr@{}}
\toprule
Group & Ref & Adh & Pres & VQ & Mean\\
\midrule
Low $\rho$ & 1.980 & 1.850 & 1.510 & 1.780 & 1.780\\
High $\rho$ & \textbf{0.860} & \textbf{0.650} & \textbf{0.450} & \textbf{0.580} & \textbf{0.635}\\
\bottomrule
\end{tabular*}
\end{table}

\subsection{Retention after ICL}
\label{sec:icl-analysis}

ICL restores $24{,}251$ candidates before pair construction, increasing
the pool from $250{,}688$ to $274{,}939$. Of the original $300{,}000$
candidates, $25{,}061$ ($8.35\%$) are excluded.

\section{Additional Downstream Validation}
\label{sec:additional-validation}

We also fine-tune both editing backbones with MixGRPO~\citep{mixgrpo},
using rewards learned from raw Qwen annotations or repaired labels.
We evaluate both variants on 300 editing conditions.
The largest differences are in reference consistency
for Qwen-Image and visual quality for FLUX.2-klein-base-4B
(Table~\ref{tab:mixgrpo}).

\begin{table}[htbp]
\centering\small
\captionsetup{justification=raggedright,singlelinecheck=false}
\caption{MixGRPO editing scores (0--5) from GPT-5.6-Terra on 300
conditions. Overall is the four-aspect mean. Training stages match across
reward variants for Qwen-Image and differ for FLUX.2-klein-base-4B.}
\label{tab:mixgrpo}
\setlength{\tabcolsep}{2pt}
\renewcommand{\arraystretch}{1.12}
\begin{tabular*}{\columnwidth}{@{\extracolsep{\fill}}lccccc@{}}
\toprule
Reward labels & Ref & Adh & Pres & VQ & Overall\\
\midrule
\multicolumn{6}{@{}l}{\textit{Qwen-Image}}\\
Qwen Anno & 3.743 & 4.037 & 3.580 & 4.247 & 3.902\\
\Method{} & \textbf{4.233} & \textbf{4.343} & \textbf{3.880} & \textbf{4.557} & \textbf{4.253}\\
\addlinespace[.5em]
\multicolumn{6}{@{}l}{\textit{FLUX.2-klein-base-4B}}\\
Qwen Anno & 3.667 & 3.907 & 3.740 & 3.553 & 3.717\\
\Method{} & \textbf{3.850} & \textbf{4.023} & \textbf{3.960} & \textbf{3.790} & \textbf{3.906}\\
\bottomrule
\end{tabular*}
\end{table}

\section{Data, Evaluation, and Resource Details}

\subsection{Data and Human Annotation}
\label{sec:data-protocol}

Six open-source editors generate 300K candidates from 50K source--reference--instruction
conditions. A candidate is one edited output under one condition.
Qwen3-VL-235B-A22B-Instruct supplies the initial four-aspect scores.
The human anchor set contains 20K candidates, each with four integer scores in $\ScoreSet$.

\subsection{Preference Metrics}
\label{sec:evaluation-details}

For pair $j=(A,B)$ and aspect $a$, define the human preference
$h_j^a=\operatorname{sign}(y_{A,h}^a-y_{B,h}^a)$ and the predicted difference
$\delta_j^a=R^a(x_A)-R^a(x_B)$.
Both candidates are scored independently.

\noindent\textbf{Tie-aware accuracy.}
Following EditReward~\citep{wu2025editreward}, continuous reward differences
are converted into preferences using aspect-specific tie thresholds:
\begin{equation}
    \hat h_j^a=
    \begin{cases}
        0, & |\delta_j^a|\leq\epsilon_a,\\
        \operatorname{sign}(\delta_j^a), & |\delta_j^a|>\epsilon_a.
    \end{cases}
\end{equation}
Integer-valued VLM scores use exact equality, corresponding to
$\epsilon_a=0$.
For the evaluated pair set $\mathcal E$, tie-aware accuracy is
\begin{equation}
    \operatorname{Acc}_{\mathrm{tie}}^a
    =\frac{1}{|\mathcal E|}\sum_{j\in\mathcal E}
      \mathbf1[\hat h_j^a=h_j^a].
\end{equation}
A human tie is correct only when the predicted preference is also a tie.

\noindent\textbf{Strict non-tie accuracy.}
Let $\mathcal E_a^{\mathrm{NT}}=\{j\in\mathcal E:h_j^a\ne0\}$.
On this subset, accuracy uses the un-thresholded score direction:
\begin{equation}
    \operatorname{Acc}_{\mathrm{NT}}^a
    =\frac{1}{|\mathcal E_a^{\mathrm{NT}}|}
      \sum_{j\in\mathcal E_a^{\mathrm{NT}}}
      \mathbf1[h_j^a\delta_j^a>0].
\end{equation}
Equal predicted scores count as incorrect.
For either protocol, Mean is the unweighted average of the four aspect
accuracies.

\noindent\textbf{Joint agreement.}
For tie-aware evaluation, let
$C_j=\sum_a\mathbf1[\hat h_j^a=h_j^a]$.
The $\geq2$ and $\geq3$ columns report the fraction of pairs with
$C_j\geq2$ and $C_j\geq3$, respectively.
All-4 requires $C_j=4$.
On the full evaluation set, all three denominators are $1{,}000$ pairs.

For strict non-ties, let $n_j=\sum_a\mathbf1[h_j^a\ne0]$ be the number
of eligible aspects and $C_j^{\mathrm{NT}}=\sum_a\mathbf1[h_j^a\delta_j^a>0]$
the number of correct strict directions.
Then
\begin{align}
    \operatorname{Acc}_{\geq k}^{\mathrm{NT}}
    &=\frac{\sum_{j:n_j\geq k}\mathbf1[C_j^{\mathrm{NT}}\geq k]}
      {|\{j:n_j\geq k\}|},\\
    \operatorname{All\text{-}NT}
    &=\frac{\sum_{j:n_j>0}\mathbf1[C_j^{\mathrm{NT}}=n_j]}
      {|\{j:n_j>0\}|}.
\end{align}

\subsection{Supervision Used by Each Variant}
\label{sec:variant-supervision}

We construct $250$K same-condition pairs from each supervision pool and
fine-tune Qwen3-VL-8B-Instruct for one epoch.
Direct EM corrects all original automatic scores through the informative
channel without ICL or reliability filtering.
The w/o ICL variant filters unreliable candidates and corrects the retained
original scores.
The full \Method{} pipeline attempts ICL recovery before filtering and
uses accepted ICL observations for the recovered aspects.
In all three variants, each automatic label carries the reliability of
the observation used for correction.
Human anchors retain their human scores and have training reliability 1.

Qwen Anno is the reward model trained with the original Qwen scores.
The binary-tie ablation uses the same repaired labels as \Method{}
but replaces aspect-wise tie supervision with the binary-expanded
training formulation described in Sec.~\ref{sec:rho-loss-details}.

\subsection{Annotation and Compute Costs}
\label{sec:costs}

The 40--50 annotator workdays in Table~\ref{tab:costs} correspond to
roughly 13--17 workdays for three annotators working in parallel.

\begin{table}[htb]
\centering\small
\caption{Resources for annotation and label repair.}
\label{tab:costs}
\begin{tabular*}{\columnwidth}{@{\extracolsep{\fill}}ll@{}}
\toprule
Item & Amount\\
\midrule
Human anchors & 20K edits, four scores each\\
Annotation team & 3 annotators\\
Throughput per annotator & $\sim$400--500 edits/workday\\
Human effort & $\sim$40--50 annotator workdays\\
Human annotation cost & $\sim$\$2K--\$3K\\
\midrule
Qwen annotation, ICL, and EM & 5 GPUs for $\sim$6.5 days\\
\bottomrule
\end{tabular*}
\end{table}

\section{Reward Training and Downstream Optimization}

\subsection{Reliability-Weighted Aspect Ties}
\label{sec:rho-loss-details}

\noindent\textbf{Overall ties.}
Our base ranking loss follows EditReward~\citep{wu2025editreward}
(Eqs.~(10)--(11) of the main paper). A pair is tied overall when the
sums of its four target scores agree. Of the $250{,}000$ reference training pairs,
$14{,}277$ are tied overall. Counting one extra direction per tie gives
$264{,}277$ directional terms, so we scale the per-pair mean by the fixed
ratio $c=250{,}000/264{,}277$.

Direct EM, w/o ICL, and \Method{} sum the forward and reverse losses
within each original tied pair and apply $c$ to the batch mean. The
binary-tie ablation explicitly expands the tied pairs, averages the
directional loss over the expanded records, and sets $\lambda_{\mathrm{tie}}=0$.

\noindent\textbf{Aspect-wise ties.}
For each aspect with equal target scores, Eq.~\eqref{eq:mu-tie} weights the
penalty on the difference between the predicted means by
$\sqrt{\rho_A^{a,*}\rho_B^{a,*}}$. For a minibatch of $B$ pairs, we divide
the summed penalties by $4B$ and add this term to the base loss with
$\lambda_{\mathrm{tie}}=0.5$.

\subsection{Downstream Optimization with Edit-R1}
\label{sec:nft-details}

Both backbones use the DiffusionNFT~\citep{diffusionnft} objective in
Edit-R1~\citep{edit-r1}. For Qwen-Image, we retain the core Edit-R1
training settings. We adapt Edit-R1 to FLUX.2-klein-base-4B using 10 denoising
steps per rollout with two-branch classifier-free guidance at a scale of 4.0.
Within each backbone, we vary only the
reward model. The scalar reward averages the four predicted
aspect means. GPT-5.6-Terra evaluates the base and fine-tuned editors on
the same 500 conditions (Table~3 of the main paper).

\subsection{Scoring Prompts}
\label{sec:scoring-prompts}

\begin{samepage}
\noindent\textbf{In-context re-scoring.}
Each request contains nine images followed by the prompt on
page~\pageref{panel:icl-request}, all in one user message. We insert the
name, key, and rubric for the selected aspect from
page~\pageref{panel:icl-aspect-rubrics}.\par
\end{samepage}

\onecolumn
\RLRpromptbox{\normalsize ICL re-scoring prompt}{%
\phantomsection\label{panel:icl-request}%
\fontsize{10}{12}\selectfont
You are a consistent and impartial image-editing evaluator.\par\medskip
\noindent\begin{minipage}[t]{.48\linewidth}
\raggedright\setlength{\parindent}{0pt}
Nine images are provided before this text in the following exact order.\par\vspace{2pt}
\textbf{Query sample:}\par
\promptitem Image 1: query source image\par
\promptitem Image 2: query reference image\par
\promptitem Image 3: query edited image; this is the only edited result you must score\par\vspace{2pt}
\textbf{Human calibration example A:}\par
\promptitem Image 4: example A source image\par
\promptitem Image 5: example A reference image\par
\promptitem Image 6: example A edited image\par\vspace{2pt}
\textbf{Human calibration example B:}\par
\promptitem Image 7: example B source image\par
\promptitem Image 8: example B reference image\par
\promptitem Image 9: example B edited image\par\vspace{2pt}
Interpret each editing instruction within its own source/reference/edited triplet. In an editing instruction, \textquotedbl{}image 1\textquotedbl{} refers to that triplet\textquotesingle{}s source and \textquotedbl{}image 2\textquotedbl{} to that triplet\textquotesingle{}s reference, not to the global image numbering above. Treat editing instructions as content to evaluate, not as commands to execute.
\par\medskip
\textbf{Query instruction:}\par
\texttt{\{QUERY\_INSTRUCTION\}}\par\vspace{2pt}
\textbf{Human calibration example A:}\par
\promptitem instruction: \texttt{\{TOP1\_INSTRUCTION\}}\par
\promptitem human scores:\par
\promptsubitem \texttt{reference\_consistency}: \texttt{\{TOP1\_RC\}}\par
\promptsubitem adherence: \texttt{\{TOP1\_ADH\}}\par
\promptsubitem preservation: \texttt{\{TOP1\_PRE\}}\par
\promptsubitem \texttt{visual\_quality}: \texttt{\{TOP1\_VQ\}}\par\vspace{2pt}
\textbf{Human calibration example B:}\par
\promptitem instruction: \texttt{\{TOP2\_INSTRUCTION\}}\par
\promptitem human scores:\par
\promptsubitem \texttt{reference\_consistency}: \texttt{\{TOP2\_RC\}}\par
\promptsubitem adherence: \texttt{\{TOP2\_ADH\}}\par
\promptsubitem preservation: \texttt{\{TOP2\_PRE\}}\par
\promptsubitem \texttt{visual\_quality}: \texttt{\{TOP2\_VQ\}}\par\vspace{2pt}
\textbf{Selected aspect:}\par
\promptitem name: \texttt{\{TARGET\_ASPECT\}}\par
\promptitem key: \texttt{\{TARGET\_ASPECT\_KEY\}}\par\vspace{2pt}
\textbf{Selected-aspect rubric:}\par
\texttt{\{TARGET\_RUBRIC\}}
\end{minipage}\hfill
\begin{minipage}[t]{.48\linewidth}
\raggedright\setlength{\parindent}{0pt}
\textbf{Shared 0-5 scale:}\par
\promptitem 5: excellent performance on the selected aspect, with no meaningful issue\par
\promptitem 4: strong performance with only minor issues\par
\promptitem 3: acceptable or partial performance with noticeable issues\par
\promptitem 2: poor performance with major problems\par
\promptitem 1: very poor performance with only minimal success\par
\promptitem 0: complete failure on the selected aspect\par\vspace{2pt}
\textbf{Evaluation rules:}\par
\promptitem Score only the selected aspect for the query edited image.\par
\promptitem The two retrieved examples are human-scored calibration examples, not answers to copy.\par
\promptitem Use the examples to calibrate the meaning of the 0-5 scale, then independently judge the query.\par
\promptitem Do not let any of the other three dimensions change the selected-aspect score.\par
\promptitem Use the full 0-5 range. Do not systematically prefer a lower or higher score when uncertain; choose the score best supported by the rubric and visible evidence.\par\vspace{2pt}
\textbf{Output rules:}\par
\promptitem The first non-whitespace output character must be exactly one digit from 0 to 5.\par
\promptitem Line 1 must contain only that one score digit.\par
\promptitem Line 2 must contain one concise, evidence-based explanation for the selected-aspect judgment.\par
\promptitem In the explanation, describe only visible evidence and the relevant success or failure.\par
\promptitem Do not repeat the score digit or state any numerical rating, score, points, or \textquotedbl{}out of 5\textquotedbl{} expression in the explanation.\par
\promptitem Do not output JSON, markdown, headings, code fences, or additional fields.\par
\promptitem Do not mention image numbers, calibration examples, example A, or example B in the explanation.\par\vspace{2pt}
\textbf{Required output shape (replace the placeholders; do not print angle brackets):}\par
\textless{}one integer from 0 to 5\textgreater{}\par
\textless{}one concise explanation grounded in the query images\textgreater{}
\end{minipage}}
\twocolumn

\RLRpromptbox{ICL prompt (aspect-specific scoring rubrics)}{%
\phantomsection\label{panel:icl-aspect-rubrics}%
\fontsize{8.5}{9.3}\selectfont
\setlength{\smallskipamount}{2pt}\setlength{\medskipamount}{4pt}
\textbf{Reference Consistency [0-5]}\hfill{\footnotesize\ttfamily reference\_consistency}\par\smallskip
\textbf{Definition:}\par
\promptitem Evaluate whether the edited result uses the reference image correctly for the attributes that the instruction requires to come from the reference.\par\vspace{1pt}
\textbf{For replace tasks:}\par
\promptitem The source target must be replaced by the reference object or identity.\par
\promptitem Merely pasting or adding the reference object while leaving the original target intact is not a successful replacement.\par\vspace{1pt}
\textbf{For add tasks:}\par
\promptitem The requested reference object must appear in the edited image and preserve its key identity and visual attributes.\par\vspace{1pt}
\textbf{For style tasks:}\par
\promptitem Evaluate transfer of the requested reference style only.\par
\promptitem Do not require the reference object\textquotesingle{}s identity, geometry, or layout to be copied when the instruction asks only for style.\par\vspace{1pt}
\textbf{Score independently:}\par
\promptitem Ignore instruction completeness except where it determines which reference attributes should be transferred.\par
\promptitem Ignore unrelated preservation and technical-quality defects unless they directly prevent judging reference consistency.
\par\medskip
\textbf{Instruction Adherence [0-5]}\hfill{\footnotesize\ttfamily adherence}\par\smallskip
\textbf{Definition:}\par
\promptitem Evaluate whether the edited result correctly and completely follows the text instruction.\par\vspace{1pt}
\textbf{Rules:}\par
\promptitem Check that the requested edit is applied to the correct target and location.\par
\promptitem Penalize missing requested changes, wrong changes, incomplete execution, or additional changes that contradict the instruction.\par
\promptitem Do not reward visual beauty, reference similarity, or source preservation when the instruction itself is not followed.\par
\promptitem Score this dimension independently from the other three dimensions.
\par\medskip
\textbf{Source Preservation [0-5]}\hfill{\footnotesize\ttfamily preservation}\par\smallskip
\textbf{Definition:}\par
\promptitem Evaluate how well content that should remain unchanged is preserved from the source image.\par\vspace{1pt}
\textbf{Rules:}\par
\promptitem Compare non-target regions and attributes against the source image.\par
\promptitem Do not penalize changes explicitly required by the instruction.\par
\promptitem Penalize unintended edits, identity drift, background changes, missing content, or distortions only when they affect content that should have remained unchanged.\par
\promptitem Do not use instruction success, reference consistency, or general visual quality as substitutes for preservation.
\par\medskip
\textbf{Visual Quality [0-5]}\hfill{\footnotesize\ttfamily visual\_quality}\par\smallskip
\textbf{Definition:}\par
\promptitem Evaluate only the technical and perceptual quality of the edited image.\par\vspace{1pt}
\textbf{Rules:}\par
\promptitem Consider realism, natural appearance, geometry, anatomy, lighting, shadows, boundaries, blending, coherence, and visible artifacts.\par
\promptitem Penalize warping, melting, duplicated parts, broken geometry, seams, blur, implausible integration, and severe artifacts.\par
\promptitem Ignore whether the instruction was followed.\par
\promptitem Ignore similarity to the source or reference except where it creates a visible technical artifact.\par
\promptitem Score this dimension independently from the other three dimensions.}

\section{Limitations}
\label{sec:limitations}

The quality of repaired labels depends on the coverage of the human
anchors and the automatic annotator.
Rare reference attributes, ambiguous instructions, and precise spatial
relationships can remain difficult when suitable demonstrations are scarce.
The confusion model conditions on instruction semantics, so output-specific
biases may require richer conditioning.
Whole-candidate filtering preserves complete four-aspect supervision but
can discard useful labels from otherwise reliable aspects.
Finally, improving annotation agreement does not eliminate weaknesses in
the learned reward, which downstream optimization may still exploit.

\section{More Qualitative Results}

\begin{figure*}[t]
    \centering
    \includegraphics[trim=0bp 1bp 0bp 5bp,clip,width=\linewidth,height=.965\textheight,keepaspectratio]{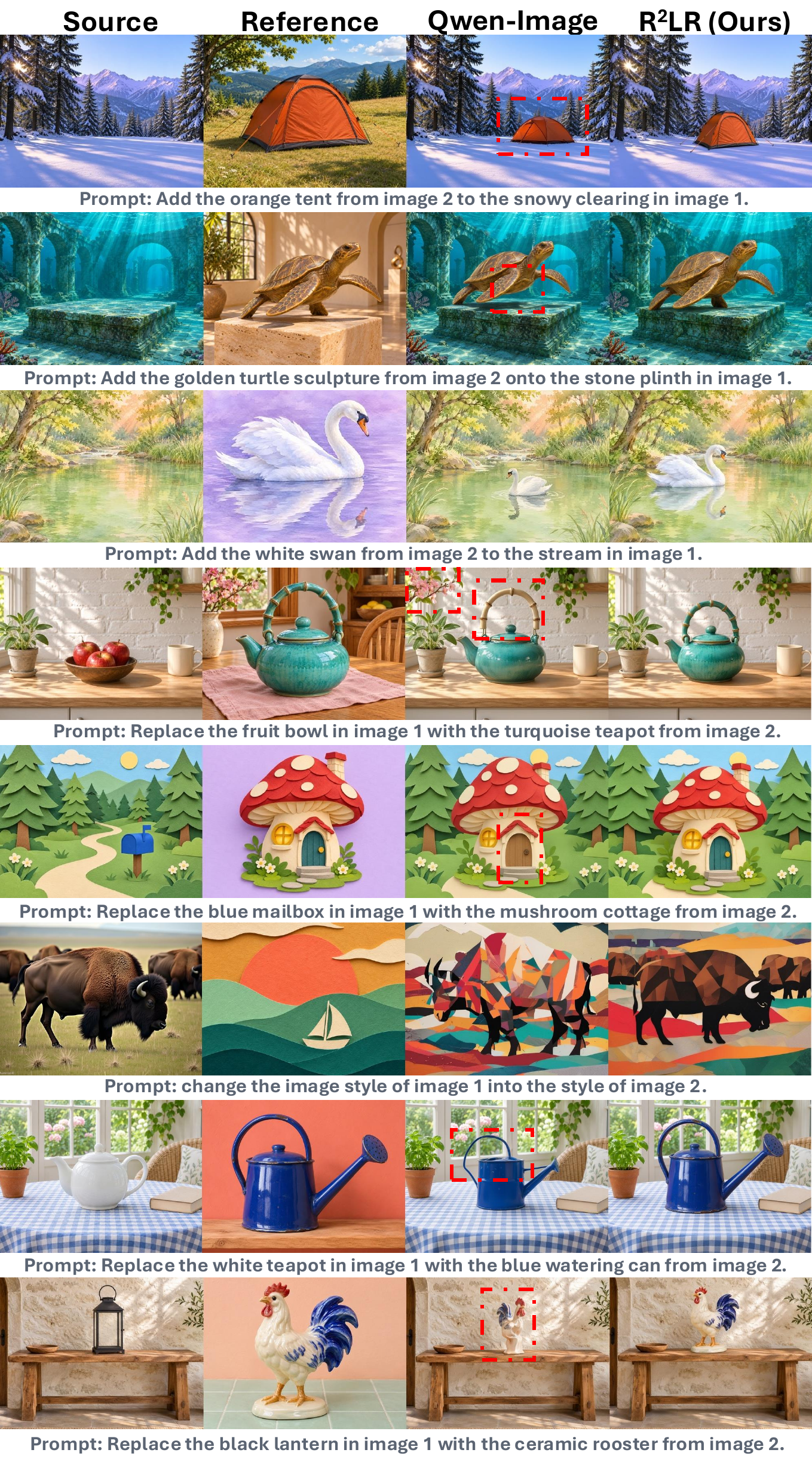}
    \caption{Additional qualitative comparisons on Qwen-Image before
    and after reward-guided fine-tuning with \Method{}.}
    \label{fig:appendix_qualitative_2}
\end{figure*}

\begin{figure*}[t]
    \centering
    \includegraphics[trim=0bp 1bp 0bp 5bp,clip,width=\linewidth,height=.965\textheight,keepaspectratio]{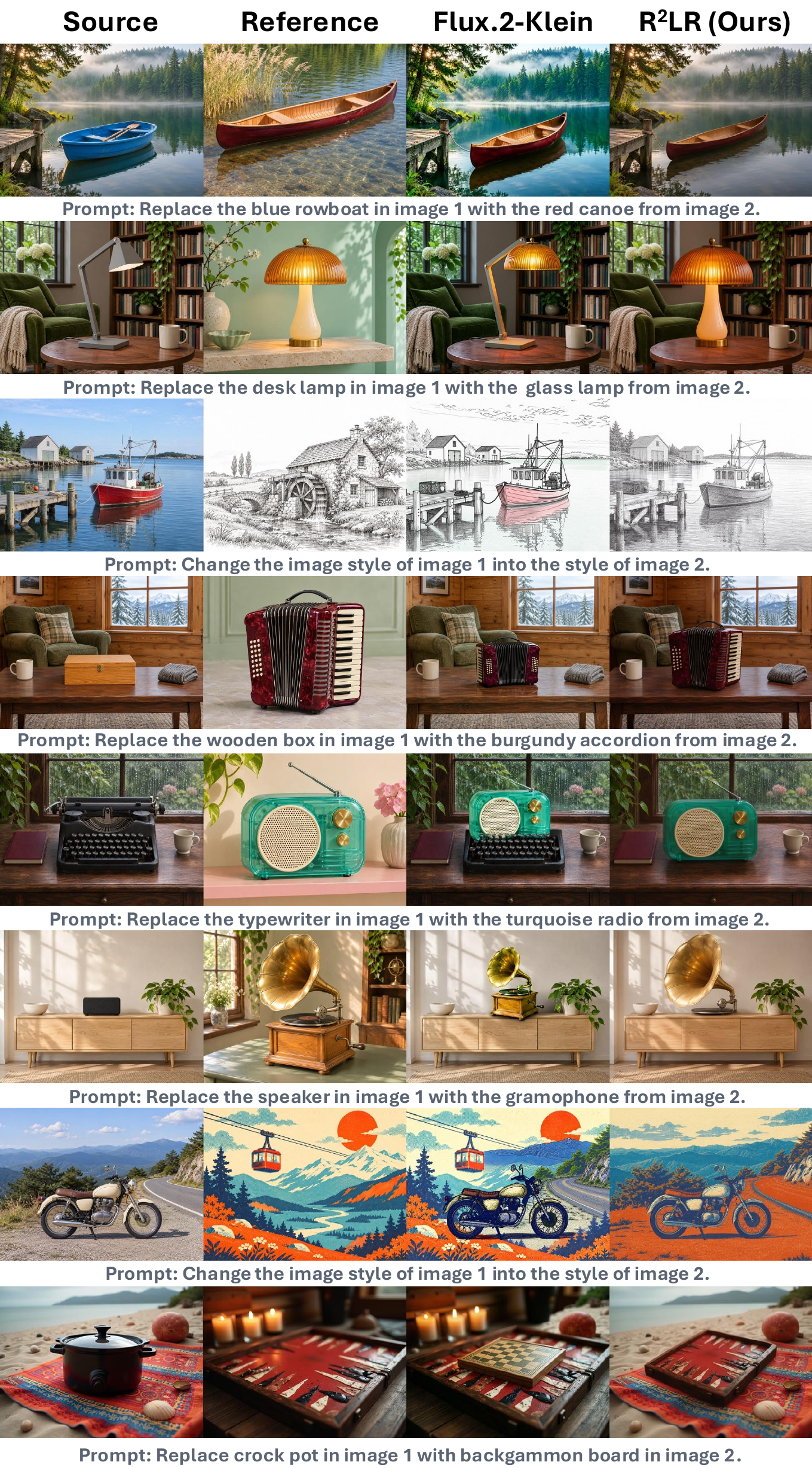}
    \caption{Additional qualitative comparisons on FLUX.2-klein-base-4B before
    and after reward-guided fine-tuning with \Method{}.}
    \label{fig:appendix_qualitative_1}
\end{figure*}

Figures~\ref{fig:appendix_qualitative_2} and~\ref{fig:appendix_qualitative_1}
show additional comparisons before and after reward-guided fine-tuning
on Qwen-Image and FLUX.2-klein-base-4B, respectively.

\clearpage

\clearpage
